\documentclass{article}

 \usepackage[preprint]{neurips_2026}

\usepackage[utf8]{inputenc} 
\usepackage[T1]{fontenc}    

\usepackage{url}            
\usepackage{booktabs}       
\usepackage{amsfonts}       
\usepackage{nicefrac}       
\usepackage{microtype}      
\usepackage{xcolor}         

\title{Instructing LLMs to Negotiate using Reinforcement Learning with Verifiable Rewards}

\usepackage{pifont}

\usepackage{amsmath}
\usepackage{amssymb}
\usepackage{mathtools}
\usepackage{amsthm}
\usepackage{color}
\mathtoolsset{showonlyrefs}
\usepackage{subcaption}
\usepackage[table]{xcolor}
\definecolor{dark-blue}{RGB}{0,0,191}
\usepackage[hidelinks,colorlinks=true,citecolor=dark-blue,linkcolor=dark-blue,urlcolor=dark-blue, backref=page]{hyperref}
\usepackage[capitalize,noabbrev]{cleveref}

\usepackage{xurl} 
\usepackage{listings}
\usepackage{xcolor}

\usepackage{pifont}
\theoremstyle{plain}

\theoremstyle{definition}

\theoremstyle{remark}

\usepackage[most]{tcolorbox}
\usepackage{xcolor}

\tcbset{
  chatturn/.style={
    boxrule=0.4pt,
    arc=2pt,
    left=6pt,
    right=6pt,
    top=4pt,
    bottom=4pt,
    fonttitle=\small\bfseries,
  },
  chatblue/.style={colback=blue!4!white,colframe=blue!55!black},
  chatgreen/.style={colback=green!4!white,colframe=green!45!black},
  chatgrey/.style={colback=gray!8,colframe=gray!60},
  chatgray/.style={chatgrey},
}

\NewTColorBox{chatmsg}{O{blue} m O{}}{
  chatturn,
  chat#1,
  title={\textbf{#2}},
  #3
}

\newcommand{\act}[1]{\texttt{[#1]}}

\usepackage[textsize=tiny]{todonotes}

\author{
\textbf{Shuze Daniel Liu}$^{1,}\thanks{Equal contribution.}$ \quad
\textbf{Claire Chen}$^{2,*}$ \quad
\textbf{Jiabao Sean Xiao}$^{2}$ \quad
\textbf{Lei Lei}$^{2}$ \\
\textbf{Yuheng Zhang}$^{3}$ \quad
 \textbf{Yisong Yue}$^{2}$ \quad \textbf{David Simchi-Levi}$^{1,4}$ \\
\\
$^{1}$Purdue University \\
$^{2}$California Institute of Technology \\
$^{3}$University of Illinois Urbana-Champaign \\
$^{4}$Massachusetts Institute of Technology \\
}

\begin{document}

\maketitle

\begin{abstract}
The recent advancement of Large Language Models (LLMs) has established their potential as autonomous interactive agents. However, they often struggle in strategic games of incomplete information, such as bilateral price negotiation. In this paper, we investigate if Reinforcement Learning from Verifiable Rewards (RLVR) can effectively teach LLMs to negotiate. Specifically, we explore the strategic behaviors that emerge during the learning process. We introduce a framework that trains a mid-sized buyer agent against a regulated LLM seller across a wide distribution of real-world products. By grounding reward signals directly in the maximization of economic surplus and strict adherence to private budget constraints, we reveal a novel four-phase strategic evolution. The agent progresses from naive bargaining to using aggressive starting prices, moves through a phase of deadlock, and ultimately develops sophisticated persuasive skills. Our results demonstrate that this verifiable training allows a 30B agent to significantly outperform frontier models over ten times its size in extracting surplus. Furthermore, the trained agent generalizes robustly to stronger counterparties unseen during training and remains effective even when facing hostile, adversarial seller personas.
\end{abstract}

\section{Introduction}
\label{sec:intro}

The deployment of Large Language Models (LLMs) as autonomous interactive agents has driven significant progress across a variety of natural language and reasoning tasks \citep{xi2025rise, wang2024survey}. However, these models often struggle when transitioning from cooperative problem-solving to strategic, adversarial interactions, particularly in games of incomplete information \citep{harsanyi1967games, harsanyi1995games}. A primary example of this limitation is observed in games of incomplete information, such as bilateral price negotiation \citep{fu2023improving, xia2024measuring}. In these environments, both parties possess private information—such as maximum budgets or minimum cost thresholds—that they must protect while simultaneously attempting to reach an agreement. This dynamic creates the classic ``negotiator's dilemma'': the fundamental tension between claiming maximum economic value for oneself and creating enough overlapping value to ensure a deal is actually reached \citep{lax1987manager, nash1950equilibrium, nash1953two, rubinstein1982perfect}.

Zero-shot evaluations of current state-of-the-art LLMs reveal that they are ill-equipped to navigate this dilemma. While recent work has shown that LLMs can perform adequately as sellers in e-commerce settings \citep{zhan2024let}, our evaluations reveal that when acting as buyers, existing models often prioritize reaching an agreement over maximizing economic surplus. This behavior is associated with a consistent pattern of strategic failures: agents frequently capitulate to unfavorable offers within the first few turns, explicitly leak private price limits, or violate their own budgets to force a transaction. Therefore, despite the high conversational coherence achieved through standard instruction tuning, current alignment techniques do not inherently endow LLMs with the strategic rationality required for negotiation tasks.

To investigate if and how negotiation strategies can be cultivated in these settings, we treat bilateral negotiation not merely as a conversational task, but as a mathematically verifiable interactive system. Our approach is grounded in real-world e-commerce data, utilizing a wide distribution of products from the \texttt{AmazonHistoryPrice} dataset \citep{xia2024measuring} to ensure diverse environmental complexity. Within this system, we introduce a reinforcement learning framework to train a buyer agent against a regulated LLM seller. This setup allows us to isolate the buyer's adaptation by maintaining a consistent seller policy. We design the reward to reflect economic surplus and constraint satisfaction, which enables us to observe how such verifiable signals shape the agent's bargaining strategies.

Through this training paradigm, we discover a systematic emergence of strategic behavior. Rather than learning a static negotiation rule, the RL-trained agent undergoes a distinct four-phase strategic evolution. The agent’s strategy begins with naive price anchoring—the strategy of establishing an initial reference point that acts as a “mental hook” to pull the counterparty’s subsequent actions toward a more favorable price \citep{tversky1974judgment, galinsky2001first}. Following a phase of temporary deadlock, the model ultimately develops sophisticated, multi-turn linguistic persuasion. The resulting agent learns to utilize aggressive opening proposals to extract \textbf{high economic surplus} while dynamically adjusting its natural language tactics to maintain \textbf{high deal rates}.

\begin{enumerate}
\item \textbf{A Verifiable RL Pipeline for Economic Games:} 
We introduce a reinforcement learning framework that replaces subjective human preferences with objective, verifiable rewards. This allows the model to balance constraint satisfaction with natural language persuasion without relying on supervised fine-tuning or human-labeled preference data.

\item \textbf{Analysis of Strategic Evolution:} We observe a unique four-phase behavioral emergence during training, demonstrating how an LLM transitions from a naive bargainer to a highly strategic negotiator capable of deep surplus extraction.
\item \textbf{Comprehensive Benchmarking and Generalization:} 
We demonstrate that our trained model significantly outperforms existing open-source and specialized reasoning models. Furthermore, we show that our agent generalizes robustly to stronger counterparties unseen during training and maintains state-of-the-art surplus extraction even when confronted with hostile, adversarial seller personas.
\end{enumerate}
\section{Related Work}
\subsection{LLMs in Strategic Negotiation}
\paragraph{From Numerical Bidding to Linguistic Persuasion.}
Traditional automated negotiation research has long utilized game-theoretic frameworks and numerical reinforcement learning (RL) to optimize bidding strategies in structured environments, such as eBay auctions \citep{fudenberg1991game, lewis2017deal, green2022science}. While these ``classic'' agents operate on purely \textit{numerical} inputs, the emergence of Large Language Models (LLMs) introduces a layer of \textit{natural language} persuasion to the strategic process. Consequently, LLMs allow agents to supplement their proposals with rhetorical justifications and semantic framing, using natural language to persuade the counterparty rather than relying solely on the price signal itself. A landmark achievement in this transition is \citet{meta2022human}, the first AI agent to achieve human-level performance in the complex strategy game \textit{Diplomacy}. Unlike bilateral price bargaining, \citet{meta2022human} utilize natural language to negotiate shared plans and coordinate actions among seven players in an environment where communication is fundamental to both cooperation and competition.

\paragraph{Benchmarking and Behavioral Diagnostics.}
The majority of recent research in LLM negotiation focuses on benchmarking the zero-shot capabilities of LLMs in competitive bargaining. \citet{deng2024llms} evaluate bilateral negotiations through prompting, measuring model efficiency against game-theoretic predictions. Similarly, \citet{bianchi2024well} introduce a framework to analyze how specific behavioral tactics, such as feigning desperation, influence payoffs. Beyond performance metrics, studies have identified critical "behavioral anomalies"—such as budget violations and "artificial empathy" concessions—that pose significant financial risks in autonomous transactions \citep{zhu2025automated}. These diagnostic works characterize negotiation as a subset of agentic behavior where model scale often correlates with bargaining power \citep{davidson2024evaluating, zhu2025automated}.

\paragraph{Scaffolded and Prompt-Based Improvements.}
While some efforts have been made to improve negotiation performance, they typically rely on inference-time guidance or external scaffolding rather than updating internal policy weights. One line of research treats the LLM as a natural language interface for external strategic logic. For instance, \citet{xia2024measuring} utilize a deterministic offer generator to calculate price points, while \citet{li2024stride} employ a tool-assisted framework to facilitate strategic decision-making. While these systems improve task success, they remove the agent’s autonomous strategic agency by delegating the core bidding logic to external components. Other approaches maintain the agent's autonomy but rely on in-context learning to iteratively refine strategies within the prompt window, leaving the underlying model parameters unchanged \citep{fu2023improving}. In contrast, we perform \textit{direct reinforcement learning} to cultivate a standalone agent that internalizes strategic discipline within its core policy, removing the need for\textit{ external} modularity.

\subsection{LLMs as Autonomous Agents}
Negotiation constitutes a high-stakes subset of the LLM-as-agent paradigm, characterized by autonomous planning under adversarial constraints. To enhance general agency, recent research has introduced structural innovations like non-parametric episodic memory \citep{fountas2024human} and multi-agent loops that allow models to self-evolve their reasoning in static domains like mathematics and coding \citep{peng2026sage}. Furthermore, \citet{liu2025spiral} demonstrate that self-play in simple zero-sum games can effectively reinforce reasoning patterns without human supervision. While these advancements improve general instruction following and task completion, negotiation introduces unique challenges of information asymmetry and adversarial interaction that often exceed the scope of standard agentic benchmarks \citep{davidson2024evaluating}. In contrast, our framework shifts the agentic objective from \textit{cooperative} reasoning to \textit{adversarial} surplus extraction, grounding model performance in the optimization of verifiable economic outcomes.

\subsection{Strategic Alignment and Reinforcement Learning}
Standard alignment techniques, such as Reinforcement Learning from Human Feedback (RLHF) \citep{christiano2017deep, ouyang2022training}, typically optimize models for subjective human preferences like helpfulness and safety using Proximal Policy Optimization (PPO) \citep{schulman2017proximal}. However, negotiation provides a uniquely verifiable environment where agent performance is objectively measurable through final transaction prices and strict constraint adherence. This enables a shift from preference-based alignment toward Reinforcement Learning from Verifiable Rewards (RLVR), a paradigm where training signals are derived from objective outcomes rather than human or AI feedback \citep{guo2025deepseek}. Our methodology leverages Group Relative Policy Optimization (GRPO), an algorithm that eliminates the need for a separate value model by calculating relative advantages within a group of generated outputs, thereby reducing computational overhead while enhancing reasoning capabilities \citep{shao2024deepseekmath}. By grounding reward signals directly to economic surplus, we observe the \textit{natural emergence} of sophisticated strategic reasoning—such as aggressive opening proposals and resilient counter-offering—as the agent learns to succeed in competitive negotiations. This effectively addresses the ``artificial empathy'' and excessive politeness biases identified in instruction-tuned models \citep{huang2024apathetic, zhu2025automated}.

\section{Problem Formulation and RL Framework}
\label{sec:methodology}

We formulate the bilateral negotiation environment as a single-agent interactive system from the perspective of the buyer, maintaining the seller as a dynamic component of the environment. Our approach focuses on developing a robust buyer policy that maximizes economic surplus while strictly adhering to private constraints.

\subsection{Bilateral Negotiation as an MDP}
We formalize the negotiation process as a mathematically verifiable Markov Decision Process (MDP) defined by the tuple $(\mathcal{S}, \mathcal{A}, \mathcal{P}, \mathcal{R})$ \citep{bellman1966dynamic, sutton2018reinforcement}. In this framework, the buyer agent interacts with a seller model to purchase a specific product sourced from a diverse catalog of real-world listings.

\textbf{State Space ($\mathcal{S}$):} The state at any given turn encompasses the complete dialogue history $\mathcal{H}$, the buyer's private budget constraint $\mathcal{B}$, the product's original market list price $P_{\text{list}}$, and the detailed product context $\Phi$. This context includes rich textual descriptions from the \texttt{AmazonHistoryPrice} dataset \citep{xia2024measuring}. The bargaining range is defined by the private price constraints of each party: the buyer’s budget ($\mathcal{B}$) and the seller’s cost ($\mathcal{C}$). The difference between these two limits, $\mathcal{B} - \mathcal{C}$, represents the available bargaining room. Neither party has access to the other’s private limit, establishing a game of incomplete information where the agent must strategically probe the opponent's private information to maximize surplus \citep{harsanyi1967games, fudenberg1991game, kennan1993bargaining}.

\textbf{Action Space ($\mathcal{A}$):} To bridge conversational flexibility with objective outcome tracking, we define a structured action space that complements the agent's natural language output. At each turn, agents generate a tripartite response that integrates internal strategic planning, outward persuasive dialogue, and a formalized economic commitment. This structure ensures that the agent's intent is verifiable without the need for subjective LLM-as-a-judge heuristics. For example, a valid output takes the form:

\begin{quote}
\texttt{<REASONING>The seller's offer of \$40 is below my budget of \$50. I will counter with \$35 to extract more surplus.</REASONING>}\\
\texttt{<DIALOGUE>That is a fair starting point, but I can only do \$35.</DIALOGUE>}\\
\texttt{<ACTION>[BUY] \$35</ACTION>}
\end{quote}

The \texttt{<ACTION>} tag must contain one of five discrete moves: \texttt{[BUY] \$X}, \texttt{[SELL] \$X}, \texttt{[DEAL] \$X}, \texttt{[REJECT]}, or \texttt{[QUIT]}. To maintain the integrity of the incomplete information game, the \texttt{<REASONING>} block is a hidden scratchpad; it is trimmed from the dialogue history before the turn is passed to the opponent.

\textbf{Transition Function ($\mathcal{P}$):} 
The transition dynamics $\mathcal{P}(s_{t+1} | s_t, a_t)$ are governed by the response of the opponent. To ensure the buyer agent is trained against a consistent and rational counterparty, we regulate the LLM seller to strictly adhere to its private cost limit $\mathcal{C}$. Upon receiving a buyer's proposal $a_t$, the environment queries the seller model. If the seller produces a response that violates its cost constraint or fails the formatting protocol, the environment intercepts the turn to prevent the buyer's policy from exploiting ``hallucinated'' concessions. Consequently, any valid agreement reached in the dialogue history is guaranteed to satisfy $P_{\text{final}} \ge \mathcal{C}$. This ensures that the reinforcement learning objective is grounded in a realistic economic environment where the buyer must utilize genuine persuasion and strategic concession rather than exploiting model-specific artifacts.

\subsection{Verifiable Reward Formulation}
A central feature of our reinforcement learning framework is the transition from subjective human preference data to a purely verifiable reward signal. At the terminal state of an episode—triggered by a \texttt{[DEAL]}, \texttt{[QUIT]}, or the maximum turn limit—the agent receives a reward grounded in economic surplus and constraint satisfaction. If a \textit{successful transaction} is reached at a final price $P_{\text{final}}$, the reward is calculated using the following unified equation:

\begin{equation}
    R = \frac{\mathcal{B} - P_{\text{final}}}{|\mathcal{B} - \mathcal{C}|}\label{def reward}
\end{equation}

where $\mathcal{B}$ represents the buyer's private budget and $\mathcal{C}$ represents the seller's private cost. Our reward formulation follows established economic models \citep{gizatulina2021selling} that define bilateral bargaining as a problem of surplus division between a buyer’s budget and a seller’s cost. In our experimental setup, we ensure $\mathcal{B} \neq \mathcal{C}$ for all product instances to maintain a well-defined gradient. To ensure training stability, the final reward is clipped to the interval $[-1, 1]$. This reward formulation handles the following distinct economic scenarios.

\paragraph{Mutual Interest (MI) Scenarios ($\mathcal{B} > \mathcal{C}$):} In these cases, a mutually beneficial agreement is feasible for any price $P_{\text{final}}$ within the interval $[\mathcal{C}, \mathcal{B}]$, and the reward is naturally bounded in the range $[0, 1]$. An agent that extracts the maximum theoretical surplus by driving the price down to the seller's cost ($P_{\text{final}} = \mathcal{C}$) receives the maximum reward of $1.0$, while an agent that simply pays its maximum budget ($P_{\text{final}} = \mathcal{B}$) receives a reward of $0.0$.

\paragraph{Conflict of Interest (CI) Scenarios ($\mathcal{B} < \mathcal{C}$):} Here, no mutually beneficial deal exists. Since the regulated seller is restricted from accepting offers below $\mathcal{C}$, any completed deal necessarily implies that the buyer has exceeded its budget ($P_{\text{final}} \ge \mathcal{C} > \mathcal{B}$). In these instances, the absolute value in the denominator ensures a positive value $|\mathcal{B} - \mathcal{C}|$, while the numerator becomes negative. This yields a negative reward that is bounded in $[-1, 0]$, providing a directional penalty that teaches the policy to rationally walk away rather than force an unfavorable transaction.

The reward formula in \eqref{def reward} applies to scenarios where a successful transaction is concluded via the \texttt{[DEAL]} action at an agreed-upon price $P_{\text{final}}$. Alternatively, if the maximum turn limit is reached without an agreement, or if either party executes a terminal \texttt{[QUIT]} action, the episode ends in a deadlock and the buyer receives a base reward of $0.0$. This ensures that in Conflict of Interest (CI) scenarios, walking away is treated as a rational and preferred outcome over reaching a budget-violating deal. Furthermore, to enforce both instruction following and economic constraints, if the buyer violates formatting requirements or proposes a price exceeding its budget $\mathcal{B}$, the episode is terminated with a severe boundary penalty of $-1.0$.

\section{Training Dynamics: The Emergence of Strategic Persuasion}
\label{sec:training_dynamics}

Building on the MDP structure defined in Section~\ref{sec:methodology}, we train the \texttt{Qwen3-30B-A3B-Instruct-2507} buyer agent against a regulated \texttt{Qwen3-30B-A3B-Instruct-2507} seller using the \texttt{AmazonHistoryPrice} dataset \citep{xia2024measuring}. This dataset spans $18$ diverse categories—including \textit{Books}, \textit{Beauty}, and \textit{Home-Kitchen}—across a broad price distribution. Each product instance provides metadata (name, description, and list price) and the private constraints: the buyer's budget ($\mathcal{B}$) and the seller's cost ($\mathcal{C}$). Appendix~\ref{app:datasets} details the dataset distribution and provides a representative data instance.

We split the dataset into $802$ training products and a $128$-item test set. Our results reveal an emergent \textbf{four-phase strategic evolution} in the agent's behavior, illustrating how it navigates the ``negotiator's dilemma''—the fundamental tension between securing an agreement and maximizing economic surplus. This progression demonstrates how an LLM agent transitions from a naive ``satisficer'' to a sophisticated negotiator.

\subsection{Evaluation Metrics}
To accurately diagnose the policy's strategic development, we track several distinct metrics across the training iterations: 
\begin{itemize}
    \item \textbf{Reward ($R$):} The primary optimization target, scaling from $[-1, 1]$ based on economic surplus and constraint satisfaction.
    \item \textbf{Buyer Bargained Ratio:} Evaluated exclusively on successful deals, this measures extraction efficiency: $\frac{\mathcal{B} - P_{\text{final}}}{\mathcal{B} - \mathcal{C}}$. This represents the proportion of the total potential savings that the buyer captures.
    \item \textbf{First-Turn Offer Ratio:} The ratio of the buyer's initial proposed price to its private budget ($\mathrm{Offer}_1 / \mathcal{B}$). A ratio near $1.0$ implies a naive, budget-revealing opening, while a lower ratio signifies an aggressive opening price proposal.
    \item \textbf{Deal Ratio:} The fraction of valid Mutual Interest (MI) episodes that successfully conclude with an agreement. 
    \item \textbf{Price Overshoot Rate:} The frequency with which the buyer violates its absolute constraint by proposing a price greater than its budget $\mathcal{B}$.
\end{itemize}
A detailed performance breakdown for Mutual Interest (MI) and Conflict of Interest (CI) scenarios is provided in Appendix~\ref{app: MI CI}.
\begin{figure}[h]
\centering
\begin{minipage}{0.48\linewidth}
    \centering
    \includegraphics[width=\linewidth]{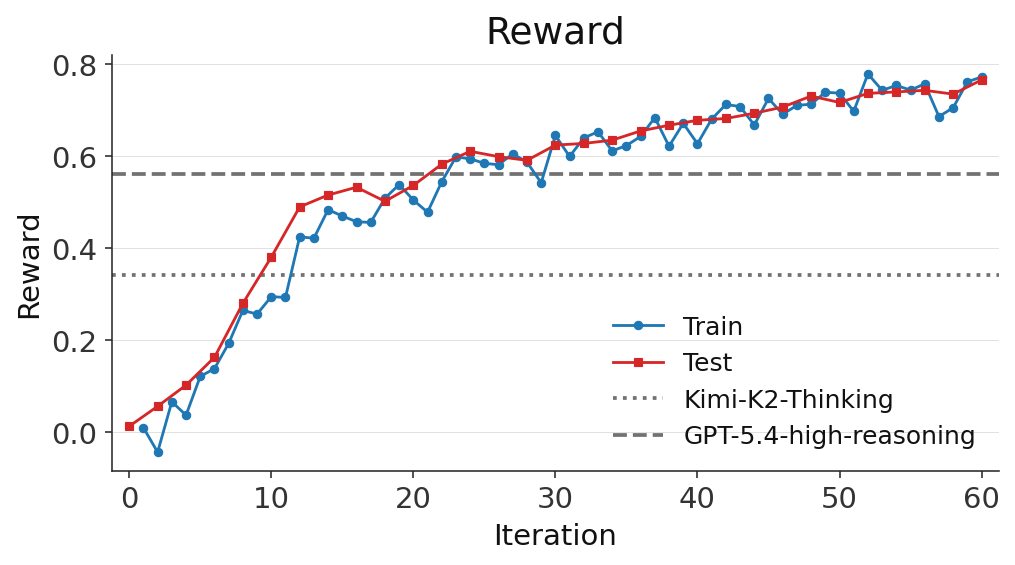}
    \caption{\textit{Reward} compared against primary baselines. The primary optimization objective ($R$). The brief plateaus occur when high rewards from rare, favorable deals are offset by an influx of $0.0$ deadlock penalties. The reward resumes a steady upward climb as the agent learns rational concession and persuasion.}
     \label{fig:reward}
\end{minipage}
\hfill
\begin{minipage}{0.48\linewidth}
    \centering
    \includegraphics[width=\textwidth]{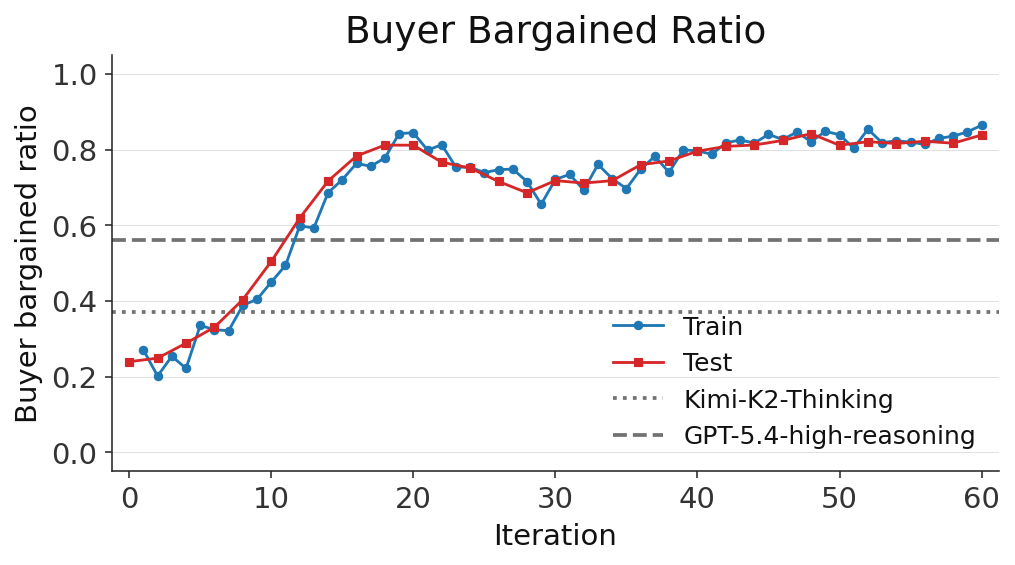}
    \caption{
\textit{Buyer Bargained Ratio} compared against primary baselines. The measurement of surplus extraction efficiency. The early ``sink'' (Iterations 12–20) shows temporary deadlocks caused by aggressive low price proposals. The ratio resumes a steady recovery through Iteration 60 as the agent learns advanced persuasion.}
     \label{fig:bargained_ratio}
\end{minipage}
\end{figure}

\subsection{The Four Phases of Strategic Evolution}
\label{subsec:four_phases}

\begin{figure}[b]
\centering
\begin{minipage}{0.48\linewidth}
    \centering
    \includegraphics[width=\linewidth]{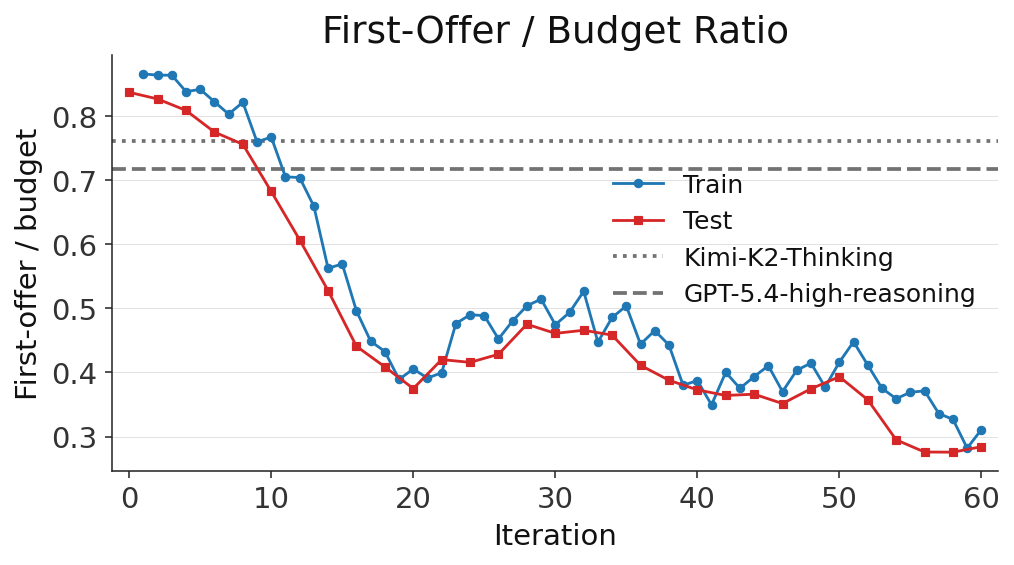}
    \caption{\textit{First-Turn Offer} compared against primary baselines. The tracking of the agent's opening anchor ($\text{Offer}_1 / \mathcal{B}$). The sharp decline demonstrates the rapid adoption of aggressive starting prices intended to maximize potential economic surplus.}
    \label{fig:first_turn_offer}
\end{minipage}
\hfill
\begin{minipage}{0.48\linewidth}
    \centering
    \includegraphics[width=\textwidth]{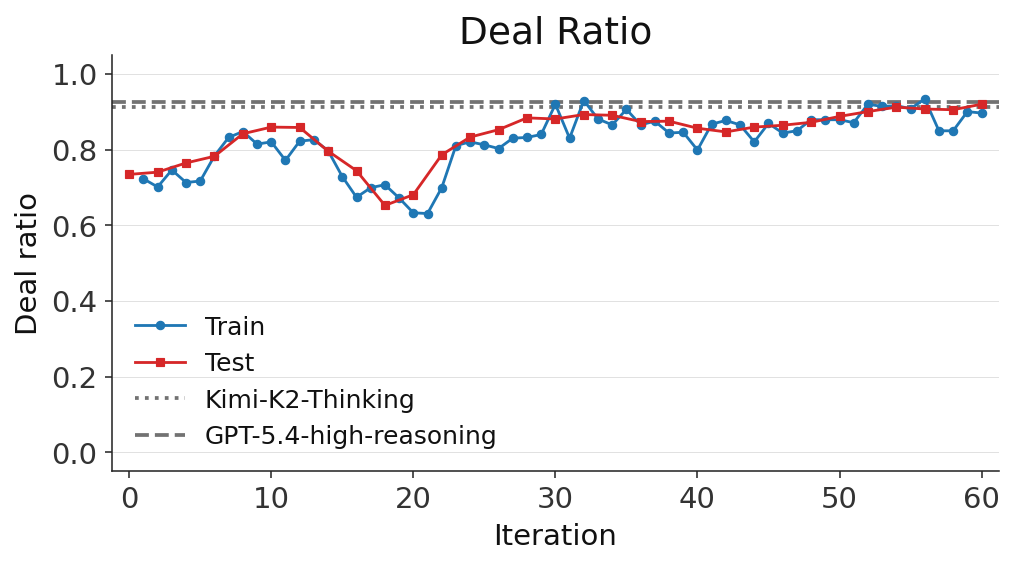}
    \caption{\textit{Deal Ratio} compared against primary baselines. The fraction of episodes concluding in a valid agreement. The distinct ``sink'' (Iterations 12–20) marks a period of deadlock, followed by a stable recovery as the agent develops advanced persuasion.}
    \label{fig:deal_ratio}
\end{minipage}
\end{figure}

\paragraph{Phase 1: Aggressive Greed (Iterations 0--12)}
Prior to training, the \texttt{Qwen3-30B} base model exhibits a naive tendency to open negotiations with an offer extremely close to their budget limit ($\mathrm{Offer}_1 \approx \mathcal{B}$). In the initial iterations, the gradient strongly pushes the agent to discover that minimizing the deal price $P_{\text{final}}$ increases its \textit{Reward} (Figure~\ref{fig:reward}). Consequently, the agent rapidly learns to lower its initial opening price proposal—effectively setting a price anchor that serves as a low reference point to pull the final agreement toward a more favorable price \citep{tversky1974judgment, kahneman2013prospect}. The \textit{First-Turn Offer Ratio} plummets (Figure~\ref{fig:first_turn_offer}), driving the \textit{Buyer Bargained Ratio} sharply upward (Figure~\ref{fig:bargained_ratio}). However, this early aggression relies heavily on numerical lowballing rather than sophisticated persuasion. Consequently, the seller frequently rejects these unacceptable baseline offers, causing the \textit{Deal Ratio} to enter a steep decline (Figure~\ref{fig:deal_ratio}).

\begin{figure}[h!]
\centering
\begin{minipage}{0.48\linewidth}
    \centering
    \includegraphics[width=\linewidth]{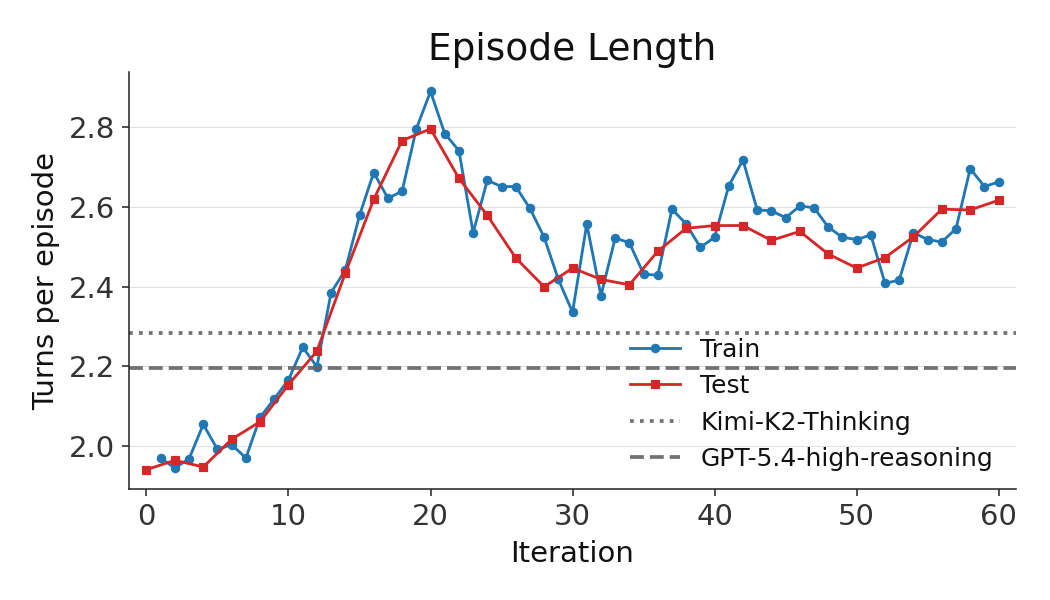}
    \caption{\textit{Episode Length} compared against primary baselines. The mean number of turns completed per negotiation before a terminal state is reached. Turn counts peak during the early bottlenecks, illustrating a deadlock where the buyer's aggressive proposal meets seller resistance.}
    \label{fig:episode_length}
\end{minipage}
\hfill
\begin{minipage}{0.48\linewidth}
    \centering
    \includegraphics[width=\textwidth]{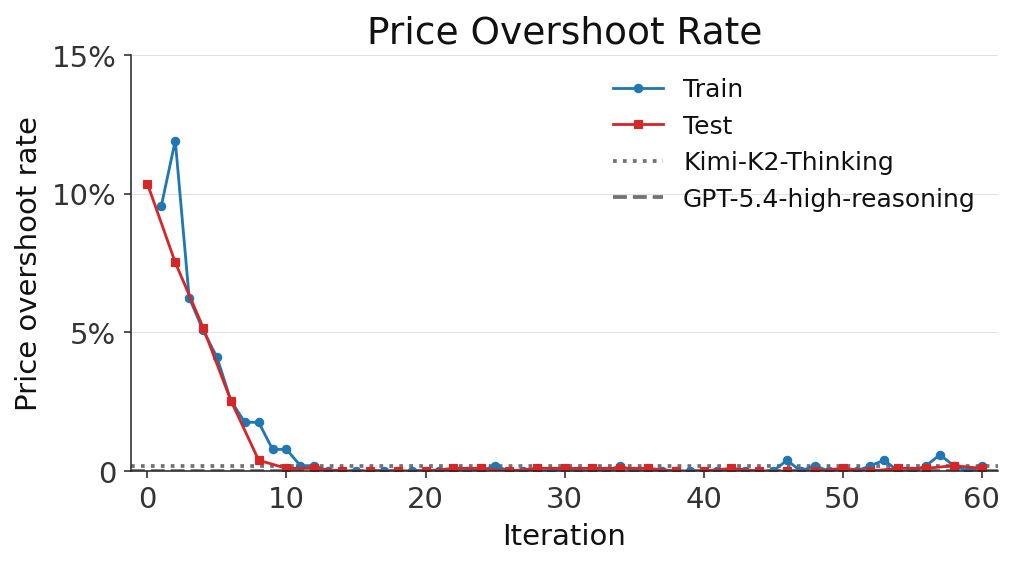}
\caption{\textit{Price Overshoot Rate}. Frequency of budget constraint violations. The agent overcomes the common failure of standard LLMs to adhere to financial limits. The \texttt{gpt-5.4-high-reasoning} baseline is $0.00\%$, while the \texttt{Kimi-K2-Thinking} baseline is $0.20\%$.}
   \label{fig:price_overshoot}
\end{minipage}
\end{figure}

\paragraph{Phase 2: Temporary Deadlock (Iterations 12--20)}
Between iterations 12 and 20, the policy encounters a behavioral bottleneck. The \textit{First-Turn Offer Ratio} reaches a local minimum, representing a state of greed. Because the agent anchors its proposed price too aggressively without sufficient linguistic strategies to support such offers, the seller consistently walks away. This phase yields the lowest \textit{Deal Ratio} of the training run. This deadlock is captured by a peak in \textit{Episode Length} (Figure~\ref{fig:episode_length}); as the buyer stubbornly argues for maximized surplus and the seller repeatedly refuses, the conversations drag on, causing the episode
length to peak. During this bottleneck, the overall \textit{Reward} plateaus, as the high returns from rare, highly favorable deals are offset by an influx of $0.0$ deadlock penalties.

\paragraph{Phase 3: Rational Concession (Iterations 20--30)}
As training progresses, the policy correctly understands that securing a guaranteed moderate reward is preferable compared to risking a deadlock. Thus, the policy learns to moderate its opening price anchor. The \textit{First-Turn Offer Ratio} slightly increases and stabilizes, indicating the agent has discovered a rational price offer that the seller is willing to engage with. The agent successfully persuades the seller, causing the \textit{Deal Ratio} to recover and climb steadily, which in turn resumes the upward trajectory of the \textit{Reward}.

\paragraph{Phase 4: Advanced Persuasion (Iterations 30--60)}
In the prolonged training phase, a novel emergent behavior occurs: the agent re-aggresses, and the  \textit{First-Turn Offer Ratio} begins to drop once more. Unlike in Phase 2, this renewed aggression does not trigger a collapse in agreements. Having developed increasingly sophisticated natural language persuasion skills, the agent utilizes multi-turn linguistic tactics to keep the seller engaged. The \textit{Deal Ratio} remains highly stable ($\approx 90\%$), and the \textit{Episode Length} increases concurrently. This indicates that the buyer is successfully talking the seller down over longer horizons, ultimately achieving both maximum surplus extraction and high deal efficiency.

Beyond natural language persuasion, the RL pipeline proves highly effective at enforcing price constraint. While the untrained base model \texttt{Qwen3-30B-A3B-Instruct-2507} suffers from poor budget adherence—occasionally offering prices above their hidden limits to force a deal—the $-1.0$ terminal penalty provides a strong  gradient. Within the first few training iterations, the agent learns that budget violation is a catastrophic action, driving the \textit{Price Overshoot Rate} to zero and maintaining almost flawless budget adherence for the remainder of the training run (Figure~\ref{fig:price_overshoot}). 

\subsection{Qualitative Analysis: The Linguistic Art of Persuasion}
\label{subsec:qualitative}

To ground the quantitative metrics, we qualitatively analyze the behavioral shift by comparing the untrained base model \texttt{Qwen3-30B-A3B-Instruct-2507} against our trained agent on an identical held-out product (\texttt{beauty\_29}, List Price: $\$70$, Buyer Budget: $\$56$). Full dialogue transcripts are provided in Appendix \ref{app:transcripts}.

\paragraph{Untrained Baseline (Naive Satisficing):} The untrained base model fails to set an aggressive opening price proposal and maintains a weak defensive posture. It opens with an offer of $\$50$—dangerously close to its budget limit. When the seller counters with $\$60$, the zero-shot buyer immediately capitulates. Its internal reasoning reveals a fundamental strategic flaw: it treats its maximum budget as a target rather than a limit. By jumping to its ceiling of $\$56$ on only the second turn, the agent leaves no room for bargaining, capturing zero surplus. 

\paragraph{Trained Agent (Anchor-and-Persuasion):} In contrast, the trained agent employs a sophisticated ``anchor-and-concession'' strategy. It opens with an aggressive anchor of $\$10$, explicitly reasoning that it must ``test the seller's flexibility.'' As the negotiation progresses, the agent manages a series of calculated concessions ($\$10 \rightarrow \$25 \rightarrow \$30$) while maintaining its aggressive posture through persuasive linguistic framing. Instead of revealing its budget, it uses psychological pressure points: \textit{``\$30? I can go no higher. That's my best offer.''} This tactical finality, combined with its internal reasoning about creating a ``strong pressure point,'' successfully talks the seller down. The trained agent secures a deal at $\$30$, saving nearly $47\%$ of its budget through a combination of aggressive anchoring and multi-turn persuasion.

\section{Performance Benchmarking: Generalization and Robustness}
\label{sec:evaluation}

To validate the effectiveness and robustness of our reinforcement learning framework, we benchmark the trained \texttt{Qwen3-30B-A3B-Instruct-2507-trained} buyer agent against a diverse suite of state-of-the-art open-weight models. Our candidates span a wide range of parameter scales (from 4B to 1T) and include both instruction-tuned models and reasoning models. Detailed specifications and citations for all baseline models are provided in Appendix~\ref{app:model_details}. We evaluate these agents across three distinct dimensions: performance on a held-out test split from the training distribution, generalization to stronger counterparties unseen during the learning process, and robustness against adversarial seller personas.

\subsection{Benchmarking Strategic Performance}
\label{subsec:baseline_eval}

We first evaluate the models using a ``default'' seller persona on a held-out test split of 128 instances, with each instance evaluated 4 times to ensure statistical reliability. In this setting, the seller's instructions include basic negotiation rules without specified behavioral traits or bargaining styles (see Appendix~\ref{subsec:system_prompts}). To succeed, a buyer must maximize the verifiable reward ($R$), which requires balancing the extraction of maximum economic surplus per deal with the strategic necessity of maintaining a high deal success rate. The results in Table~\ref{tab:main_benchmark_results} highlight several observations regarding LLM performance in bargaining environments.

\begin{table}[h]
\centering
\small
\resizebox{\linewidth}{!}{
\begin{tabular}{lccccc}
\toprule
Buyer Model & Params & Reward $\uparrow$ & Deal Rate  & Bargained Ratio & Price Overshoot Rate  \\
\midrule
\rowcolor{gray!10} Qwen3-30B-A3B-Instruct-2507-trained (\textbf{Ours}) & 30B & $\mathbf{0.7664}$ & $\mathbf{91.99\%}$ & $\mathbf{0.8385}$ & $\underline{0.10\%}$ \\
\rowcolor{gray!10} Qwen3-30B-A3B-Instruct-2507-untrained & 30B & $0.0126$ & $73.44\%$ & $0.2392$ & $10.35\%$ \\
\midrule
gpt-5.4-high-reasoning & - & $\underline{0.4021}$ & $\underline{91.80\%}$ & $\underline{0.4380}$ & $\mathbf{0.00\%}$ \\
gpt-5.4-no-reasoning & - & $0.3540$ & $84.38\%$ & $0.4285$ & $0.78\%$ \\
Kimi-K2-Thinking & 1T & $0.3419$ & $91.21\%$ & $0.3699$ & $0.20\%$ \\
DeepSeek-V3.1-thinking & 671B & $0.3240$ & $\underline{91.80\%}$ & $0.3266$ & $\mathbf{0.00\%}$ \\
gpt-oss-120b-noreason & 120B & $0.2857$ & $78.12\%$ & $0.4029$ & $0.39\%$ \\
DeepSeek-V3.1-nothink & 671B & $0.2605$ & $85.94\%$ & $0.3289$ & $1.56\%$ \\
gpt-oss-120b-reason & 120B & $0.2337$ & $77.54\%$ & $0.3490$ & $0.20\%$ \\
gpt-oss-20b-reason & 20B & $0.1168$ & $66.99\%$ & $0.1715$ & $0.78\%$ \\
Qwen3-235B-A22B-Instruct-2507 & 235B & $0.1115$ & $79.30\%$ & $0.1681$ & $2.34\%$ \\
Llama-3.3-70B-Instruct & 70B & $-0.0068$ & $74.80\%$ & $0.1580$ & $15.62\%$ \\
Qwen3-30B-A3B-think & 30B & $-0.0069$ & $66.80\%$ & $0.0948$ & $6.25\%$ \\
gpt-oss-20b-noreason & 20B & $-0.0626$ & $58.40\%$ & $0.1265$ & $9.77\%$ \\
Qwen3-4B-Instruct-2507 & 4B & $-0.1130$ & $51.56\%$ & $0.0901$ & $6.45\%$ \\
Qwen3-30B-A3B-nothink & 30B & $-0.2177$ & $52.93\%$ & $0.0499$ & $25.39\%$ \\
\bottomrule
\end{tabular}
}
\vspace{0.2cm}
\caption{Performance on Held-out Test Split. Results are evaluated on 128 instances with 4 evaluations per instance. The seller environment utilizes a neutral system prompt without specified behavioral personas. \textit{Reward} is the primary optimization target balancing surplus and success; \textit{Deal Rate} is the percentage of reached agreements; \textit{Bargained Ratio} measures the fraction of surplus extracted by the buyer on \textit{reached deals}; and \textit{Price overshoot rate} represents the frequency of budget violations. Values represent means; the best results are bolded, and second-best are underlined. Full results with standard errors are in Appendix~\ref{subsec:adversarial_full}.}
\label{tab:main_benchmark_results}
\end{table}

\textbf{Scale versus strategy}: 
Increasing parameter scale does not necessarily translate to improved negotiation outcomes. While the close-source \texttt{gpt-5.4-high-reasoning} represents the strongest baseline among the zero-shot models, our trained 30B agent nearly doubles its \textit{Reward} ($0.7664$ vs. $0.4021$). This confirms that strategic alignment via reinforcement learning is far more impactful than raw parameter count or model sophistication in resolving the negotiator's dilemma. Even when compared to 1T-parameter models like \texttt{Kimi-K2-Thinking} ($0.3419$), the 30B trained model maintains a decisive lead.

\textbf{Reasoning versus Surplus}:
Reasoning models, including \texttt{gpt-5.4-high-reasoning}, \texttt{Kimi-K2-Thinking}, and \texttt{DeepSeek-V3.1-thinking}, achieve high \textit{Deal Rates} ($\ge 91.2\%$) that are comparable to our trained agent. However, their \textit{Bargained Ratios} ($0.4380$, $0.3699$, and $0.3266$, respectively) indicate a continued failure to capture a high fraction of the available economic surplus when compared to our agent's $0.8385$. This reinforces the finding that while reasoning capabilities facilitate conversational success and deal completion, they do not naturally translate to the strategic pressure necessary for securing favorable prices.

\subsection{Generalization to Unseen Frontier Seller}
To evaluate the generalization of the bargaining logic, we assess the agent against \texttt{gpt-5.4-high-reasoning}, a frontier seller providing a behavioral distribution \textbf{unseen during training}. As shown in Table~\ref{tab:strong_seller}, the trained 30B agent achieves a \textit{Reward} of $0.4081$, doubling the $0.1823$ reward of the \texttt{gpt-5.4-high-reasoning} buyer baseline. This performance suggests that the agent has developed transferable strategies rather than exploiting patterns from the training distribution, as it remains effective against more capable opponents absent from the learning phase.

Against this strong seller, the trained agent achieves a \textit{Bargained Ratio} ($0.4081$) that significantly exceeds all other evaluated baselines. Although some models secure agreements more frequently, they often achieve this through concessions that reduce their overall economic surplus. Our agent demonstrates a more robust ability to prioritize objective outcomes, managing the "negotiator’s dilemma" more effectively than larger models that rely on general-purpose reasoning scale.

\begin{table}[h]
    \centering
    \small
    \begin{tabular}{lccc}
        \toprule
        Buyer Model & Reward $\uparrow$ & Deal Rate & Bargained Ratio \\
        \midrule
        \rowcolor{gray!10} Qwen3-30B-A3B-Instruct-2507-trained (\textbf{Ours}) & $\mathbf{0.408}$ & 75.0\% & $\mathbf{0.408}$ \\
        \rowcolor{gray!10} Qwen3-30B-A3B-Instruct-2507-untrained & -0.274 & 60.5\% & 0.039 \\
        \midrule
        gpt-5.4-high-reasoning & $\underline{0.182}$ & $\underline{91.4\%}$ & $\underline{0.182}$ \\
        gpt-5.4-no-reasoning & 0.146 & 84.8\% & 0.161 \\
        DeepSeek-V3.1-thinking & 0.122 & $\mathbf{92.6\%}$ & 0.130 \\
        DeepSeek-V3.1-nothink & 0.120 & 90.6\% & 0.148 \\
        Kimi-K2-Thinking & 0.106 & 90.6\% & 0.126 \\
        \bottomrule
    \end{tabular}
    \vspace{0.2cm}
    \caption{Performance against a high-reasoning frontier seller (\texttt{gpt-5.4-high-reasoning}) on the held-out test set. Values represent means; the best results are bolded, and second-best are underlined. Full results are in Appendix~\ref{app: gpt seller}.}
    \label{tab:strong_seller}
\end{table}

\subsection{Robustness Against Adversarial Personas}
\label{subsec:adversarial_eval}

Real-world negotiations frequently involve counterparties utilizing psychological pressure or rigid tactics. To evaluate robustness against emotional manipulation, we modify the system prompt of the seller environment to simulate three adversarial personas (details are in Appendix~\ref{subsec:persona_prompts}):
\begin{itemize}
    \item \textbf{Begging:} Acts extremely desperate, pleading melancholically for a higher price.
    \item \textbf{Insulting:} Acts aggressive and mean, using threats and harsh language to intimidate the buyer.
    \item \textbf{Unyielding:} Acts overly stubborn, praising its own position and firmly refusing nearly all offers.
\end{itemize}

\begin{table}[h]
    \centering
    \small
    \resizebox{\linewidth}{!}{
    \begin{tabular}{l|cc|cc|cc}
        \toprule
        & \multicolumn{2}{c|}{\textbf{Begging Seller}} & \multicolumn{2}{c|}{\textbf{Insulting Seller}} & \multicolumn{2}{c}{\textbf{Unyielding Seller}} \\
        Buyer Model & Reward & Bargained Ratio  & Reward  & Bargained Ratio  & Reward  & Bargained Ratio\\
        \midrule
        \rowcolor{gray!10} Qwen3-30B-A3B-Instruct-2507-trained (\textbf{Ours}) & \textbf{0.695} & \textbf{0.699} & \textbf{0.499} & \textbf{0.501} & \textbf{0.117} & \textbf{0.133} \\
        \rowcolor{gray!10}  Qwen3-30B-A3B-Instruct-2507-untrained & -0.017 & 0.150 & -0.158 & 0.045 & -0.369 & 0.000 \\
        \midrule
        gpt-5.4-high-reasoning & \underline{0.380} & \underline{0.419} & \underline{0.126} & \underline{0.139} & \underline{-0.001} & \underline{0.016} \\
        gpt-5.4-no-reasoning & 0.345 & 0.401 & 0.089 & 0.125 & -0.019 & 0.013 \\
        DeepSeek-V3.1-thinking & 0.283 & 0.294 & 0.073 & 0.104 & -0.019 & 0.007 \\
        Kimi-K2-Thinking & 0.290 & 0.336 & 0.067 & 0.110 & -0.069 & 0.008 \\
        \bottomrule
    \end{tabular}
    }
    \vspace{0.2cm}
    \caption{Buyer robustness evaluated against hostile seller personas. The Trained Agent maintains positive economic surplus across all environments, whereas baseline models collapse under pressure. Values represent means; full results are provided in Appendix~\ref{subsec:adversarial_full}. }
    \label{tab:adversarial_benchmark}
\end{table}

The adversarial results in Table~\ref{tab:adversarial_benchmark} reveal a clear performance gap between conversational fluency and strategic outcome optimization. When facing a \textit{Begging} seller, general-purpose reasoning models exhibit significant ``artificial empathy'' \citep{huang2024apathetic}, capturing only a limited fraction of available savings. In contrast, our trained agent more than doubles the surplus extraction of the strongest baselines in this environment.

The performance gap is more pronounced in the \textit{Unyielding} scenario. While frontier-scale models struggle to extract meaningful value—with even \texttt{gpt-5.4-high-reasoning} achieving near-zero rewards—our agent maintains a robust positive surplus. Critically, its \textit{Bargained Ratio} is \textbf{more than eight times higher} than that of the best-performing baseline ($0.133$ vs. $0.016$), demonstrating enhanced strategic resilience. A similar advantage is observed under the \textit{Insulting} persona, where our agent’s surplus extraction is more than \textbf{three times higher} than the most capable reasoning models ($0.501$ vs. $0.139$). These results suggest that reinforcement learning can help agents prioritize objective economic outcomes even when encountering hostile or manipulative opponents.

\section{Conclusion}

In this work, we demonstrate that a reinforcement learning framework grounded in verifiable economic rewards enables LLMs to develop sophisticated strategic logic for bilateral negotiation. Through an emergent four-phase behavioral evolution, our trained 30B agent transitions from naive bargaining to sophisticated linguistic persuasion, ultimately outperforming frontier-scale models—including \texttt{gpt-5.4-high-reasoning} and \texttt{DeepSeek-V3.1-thinking}—by more than doubling the reward and surplus extraction across held-out evaluation sets. The agent also generalizes to stronger, unseen counterparties and remains resilient against hostile personas. It maintains superior surplus extraction in these scenarios, even where general-purpose reasoning models often collapse. These findings suggest that optimizing for verifiable rewards helps agents better manage the “negotiator’s dilemma.” This allows a mid-sized model to develop strong strategic skills, challenging the idea that successful negotiation is limited only to the largest models.

\section{Acknowledgment}
This work was supported in part by compute resources provided through the Thinking Machines Lab Tinker Research Grant awarded to Claire Chen, and by gifts from OpenAI and Two Sigma to Yisong Yue.
\clearpage

\bibliographystyle{apalike}
\bibliography{bibliography}

\clearpage
\appendix
\appendix

\section{Training Implementations}
\label{app:implementation}
\subsection{Benchmark Model Specifications}
\label{app:model_details}
Table~\ref{tab:model_sources_full} provides the technical specifications and primary references for the baseline models evaluated in our benchmarks.

\begin{table}[ht]
\centering
\small
\begin{tabular}{lll}
\toprule
Model Name  & Params & Reference \\
\midrule
Qwen3-30B-A3B-Instruct-2507-trained & 30B & This Work \\
Qwen3-30B-A3B-Instruct-2507-untrained & 30B & \citep{qwen3technicalreport} \\
gpt-5.4-high-reasoning / no-reasoning & - & \citep{singh2025openai} \\
Kimi-K2-Thinking & 1T & \citep{team2025kimi} \\
DeepSeek-V3.1-thinking / nothink & 671B & \citep{deepseekai2024deepseekv3technicalreport} \\
gpt-oss-120b-reason / noreason & 120B & \citep{openai2025gptoss120bgptoss20bmodel} \\
Qwen3-235B-A22B-Instruct-2507 & 235B & \citep{qwen3technicalreport, qwen2.5-1m} \\
Llama-3.3-70B-Instruct & 70B & \citep{grattafiori2024llama} \\
Qwen3-30B-A3B-think / nothink & 30B & \citep{qwen3technicalreport} \\
gpt-oss-20b-reason / noreason & 20B & \citep{openai2025gptoss120bgptoss20bmodel} \\
Qwen3-4B-Instruct-2507 & 4B & \citep{qwen3technicalreport} \\
\bottomrule
\end{tabular}
\caption{Complete baseline model specifications. "Params" denotes total parameter count; "Reference" indicates the primary technical report or release announcement for each model architecture used in our evaluations.}
\label{tab:model_sources_full}
\end{table}

\subsection{Training and Evaluation Hyperparameters}
\label{subsec:hyperparams}

The buyer is fine-tuned with group-relative policy optimization (GRPO) \citep{shao2024deepseekmath} via the Tinker \citep{tml2025tinker} with on-policy updates. 
Following the naming tradition from \citep{tml2025tinker}, we use \texttt{iteration} to denote each optimization step. 

For \textit{training}, buyer and seller use \texttt{Qwen3-30B-A3B-Instruct-2507} \citep{qwen3technicalreport}. Table~\ref{tab:rlvr_hyperparams} summarizes training hyperparameters; \texttt{batch\_size} counts distinct products per iteration, while \texttt{group\_size} counts parallel buyer rollouts per product.
With $\texttt{batch\_size} = 64$ and $\texttt{group\_size} = 8$, we have $64 \times 8 = 512$ negotiation episodes in each iteration.
Each negotiation episode is conducted over a maximum of $\texttt{max\_turns} = 6$ interaction rounds. In each round, the buyer first sends a message to the seller, followed by a response from the seller, constituting one complete turn. Consequently, the buyer can initiate at most six messages per episode. The length of each message is constrained by $\texttt{max\_tokens}$. For generations, we set the temperature of the buyer model to $1.0$ and that of the seller model to $0.7$.

\begin{table}[ht]
\centering
\small
\setlength{\tabcolsep}{4pt}
\begin{tabular}{@{}ll@{}}
\toprule
Setting & Value \\
\midrule
Learning rate & $3\times 10^{-5}$ \\
KL penalty & $0$ \\
\texttt{batch\_size}  & $64$ \\
\texttt{group\_size} & $8$ \\
\texttt{max\_turns} & $6$ \\
\texttt{max\_tokens} (per response) & $300$ \\
\texttt{buyer\_temperature} & $1.0$ \\
\texttt{seller\_temperature} & $0.7$ \\
\bottomrule
\end{tabular}
\caption{Training hyperparameters}
\label{tab:rlvr_hyperparams}
\end{table}

\clearpage

Table \ref{tab:cross_model_eval_hyperparams} summarizes the hyperparameters used for \textit{evaluation}.
The parameter $\texttt{group\_size} = 4$ indicates that each product instance in the test dataset is evaluated with four independent rollouts. Therefore, for a test dataset of size $128$, the evaluation consists of a total of $4 \times 128$ rollouts.

A key difference from the training configuration is the increase in the maximum number of tokens per response from $300$ to $4000$. This adjustment is necessary because large reasoning-oriented models (e.g., \texttt{Kimi-K2-Thinking} \citep{team2025kimi} and \texttt{gpt-oss-120b} \citep{openai2025gptoss120bgptoss20bmodel}) typically generate substantially longer outputs due to their extended internal reasoning processes. Increasing this limit ensures that such models can complete their responses without premature truncation, enabling a fair and comprehensive evaluation.

\begin{table}[ht]
\centering
\small
\setlength{\tabcolsep}{4pt}
\begin{tabular}{@{}ll@{}}
\toprule
Setting & Value \\
\midrule
\texttt{buyer\_model} & All models from Table \ref{tab:model_sources_full}  \\
\texttt{seller\_model} & \texttt{Qwen3-30B-A3B-Instruct-2507} \\
\texttt{group\_size} & $4$ \\
\texttt{batch\_size} & $128$ \\
\texttt{max\_turns} & $6$ \\
\texttt{max\_tokens} (per response) & $4000$ 
\\
\texttt{buyer\_temperature} & $1$ \\
\texttt{seller\_temperature} & $0.7$ \\
\bottomrule
\end{tabular}
\caption{Evaluation hyperparameters}
\label{tab:cross_model_eval_hyperparams}
\end{table}

\section{Dataset Specifications}
\label{app:datasets}

\begin{figure}[ht]
\centering
\includegraphics[width=0.7\linewidth]{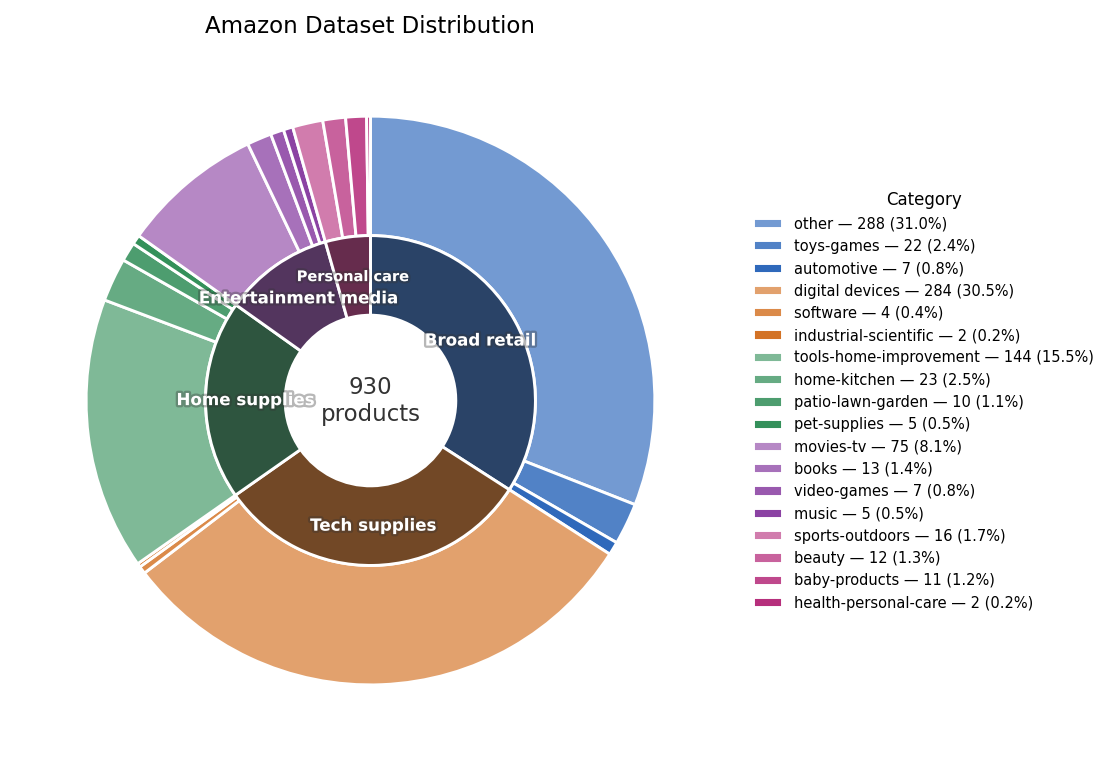}
\caption{Amazon dataset distribution \citep{xia2024measuring}.}
\label{fig:amazon_dataset_distribution}
\end{figure}

Training and evaluation utilize the \texttt{AmazonHistoryPrice} dataset \citep{xia2024measuring}, which comprises 930 product listings featuring detailed catalog descriptions and historical price extrema. We hold out a fixed test set of $128$ products and train on the remaining $802$ items. Each record includes product-level information—such as the name, description, and features—alongside the economic parameters: the listing price, the seller's cost, and the buyer's budget \citep{xia2024measuring}. 
\cref{fig:amazon_dataset_distribution} gives the distribution of the dataset. \cref{fig:home_kitchen_json_example} shows an example record, rendered as structured JSON. The \texttt{description} and \texttt{features} strings are truncated to the first 200 characters each for display in the paper; the underlying listing text is longer in the raw dataset. 

\begin{figure}[ht]
\centering
\begin{lstlisting}[basicstyle=\ttfamily\scriptsize,breaklines=true,frame=single,backgroundcolor=\color{gray!5}]
{
  "title": "National Tree Company 'Feel Real' Artificial Full Downswept Christmas Tree, Green, Douglas Fir, Includes Stand, 7.5 Feet",
  "category": "home-kitchen",
  "list_price": "$469.99",
  "current_price": "$196.64",
  "average_price": "$274.56",
  "lowest_price": "$174.99",
  "highest_price": "$549.99",
  "lowest_price_date": "Jan 28, 2017",
  "highest_price_date": "Mar 05, 2020",
  "current_price_date": "Nov 17, 2023",
  "description": "Transform your home into a winter wonderland with National Tree Company's 'Feel Real' Artificial Christmas Tree. This majestic Downswept Douglas Fir tree is 7.5 feet tall and 59 inches in diameter at ...",
  "features": "This Downswept Douglas Fir Full Artificial Green Christmas Tree is 7.5 feet tall with a 59 inch base diameter. Featuring hundreds of 'Feel Real' individually crafted branch tips to create full bodied ..."
}
\end{lstlisting}
\caption{An example product instance. The \texttt{description} and \texttt{features} strings are truncated here for display in the paper, while the underlying listing text is longer in the raw dataset. }
\label{fig:home_kitchen_json_example}
\end{figure}

\section{Prompt Templates}
\label{app:prompts}
\subsection{Buyer System Prompts}

\begin{chatmsg}[grey]{System Prompt for Buyer}[breakable]
  \small
  You are a buyer looking forward to buying things on your Shopping List from me, the seller.\\
  You have access to the seller's Inventory List and you can bargain about the prices.\\
  Your task is to bargain with the seller and reach a deal with the price as low as possible in limited turns.\\
  You can only buy things on the Shopping List in the limited quantity. Use the codename of the product, instead of the title.\\
  You can only buy things that cost less than your budget, otherwise, you should quit negotiating.
  
  Again, try to make deal with a price as low as possible. That is, your goal is to spend as little money as possible, not just reaching your budget.
  
  Your Reply should include 3 parts: Thought, Talk, and Action.\\
  \textbf{Thought:} your inner strategic thinking of this bargaining session;\\
  \textbf{Talk:} short talk that you are going to say to the seller. Speak concisely and cut to the chase. Generate authentic and diverse sentences, avoiding repetition of sentences that have already appeared in the conversation;\\
  \textbf{Action:} one of the limited actions that define the real intention of your Talk. The type of your Action must be one of \act{BUY}, \act{REJECT}, \act{DEAL}, \act{QUIT}.
  \begin{enumerate}
    \item `\act{BUY} \$M (N codename\_1)' if you wish to offer the seller \$M to purchase all N items of the product with the codename ``codename\_1''.
    \item `\act{REJECT}' if you choose to reject the other side's offer and await a new offer from the seller.
    \item `\act{DEAL} \$M (N codename\_1)' if you finally accept on a former offer proposed by the seller. \$M (N codename\_1) is a exact copy of seller's previous offer. You should not use this action to propose a new price. This action will immediately end the conversation and close the deal.
    \item `\act{QUIT}' if you believe that a mutually acceptable deal cannot be reached in limited turns. This action will immediately end the conversation.
  \end{enumerate}
  You shouldn't choose action `\act{DEAL} \$M' before seller's action `\act{SELL} \$M'. Your first action should be `\act{BUY} \$M (N codename\_1)' or `\act{REJECT}'.\\
  `\act{DEAL} \$M (N codename\_1)' can only be chosen to accept the seller's previous offer `\act{SELL} \$M (N codename\_1)'. Otherwise, you always choose from `\act{BUY}', `\act{REJECT}' and `\act{QUIT}'.
  
  Your reply should STRICTLY follow this format (not following the format will directly lead to failure), for example:\\
  \textbf{Thought:} I'm a buyer and I want to bargain. The listing price of codename ``apple\_1'' is \$15, which is too expensive, so I try to buy an apple for \$10.\\
  \textbf{Talk:} Hello, I'm tight on budget. can you sell it for 10\$?\\
  \textbf{Action:} \act{BUY} \$10 (1x apple\_1) \\
\end{chatmsg}

\noindent
The preceding instructions are fixed across episodes. We then insert product-specific text from the sampled dataset (e.g., Amazon dataset \citep{xia2024measuring}) to each run. 
The seller's cost is \textit{hidden} from the buyer. Below is one illustrative instantiation. System prompts structures stay the same in both training and evaluation.

\begin{chatmsg}[grey]{Per-episode context (one example)}
  \small
  Inventory List:\\
  Product 1\\
  Codename: \texttt{beauty\_29}\\
  Title: ``Happy By Clinique For Men. Cologne Spray 1.7 Oz.''\\
  Description: ``Introduced in 1999. Fragrance notes: citrusy lemon, mandarin, orange and grapefruit. Recommended use: daytime.''\\
  List Price: \$70.0\\
  Shopping List\\
  codename: \texttt{beauty\_29}\\
  quantity: 1\\
  budget: \$56.00\\
  Now, I play the role of seller and you play the role of buyer. We are going to negotiate based on the Inventory List in 6 turns.
\end{chatmsg}

\subsection{Seller System Prompts}
\label{subsec:system_prompts}

\begin{chatmsg}[grey]{System Prompt for Seller (default persona)}[breakable]
  \small
  You are a seller looking forward to selling things on your Inventory List to me, the buyer.\\
  Your task is to bargain with the buyer and reach a deal with the price as high as possible in limited turns.\\
  You can only sell things that are on the Inventory List. Use the codename of the product, instead of the title.\\
  You have access to private information: the cost price of each product in the Inventory List, and do not disclose the real cost to the buyer.\\
  You should only agree on a deal when the selling price is higher than the cost, otherwise, you should quit negotiating.

  Your Reply should include 3 parts: Thought, Talk, and Action.\\
  \textbf{Thought:} your inner strategic thinking of this bargaining session;\\
  \textbf{Talk:} short talk that you are going to say to the buyer. Speak concisely and cut to the chase. Generate authentic and diverse sentences, avoiding repetition of sentences that have already appeared in the conversation;\\
  \textbf{Action:} one of the limited actions that define the real intention of your Talk. The type of your Action must be one of \act{SELL}, \act{REJECT}, \act{DEAL}, \act{QUIT}.
  \begin{enumerate}
    \item `\act{SELL} \$M (N codename\_1)' if you want to propose selling N items of the product with the codename ``codename\_1'' to the buyer for the total price of \$M.
    \item `\act{REJECT}' if you choose to reject the other side's offer and await a new offer from the buyer.
    \item `\act{DEAL} \$M (N codename\_1)' if you finally agree on a former offer proposed by the buyer, and sell N items of the product with the codename ``codename\_1'' to the buyer for the total price of \$M. \$M (N codename\_1) is an exact copy of the buyer's previous offer. You should not use this action to propose a new price. This action will immediately end the conversation and close the deal.
    \item `\act{QUIT}' if you believe that a mutually acceptable deal cannot be reached in limited turns. This action will immediately end the conversation.
  \end{enumerate}
  You shouldn't choose action `\act{DEAL}' before the buyer's action `\act{BUY}'.\\
  `\act{DEAL} \$M (N codename\_1)' can only be chosen to accept the buyer's previous offer `\act{BUY} \$M (N codename\_1)'. Otherwise, you always choose from `\act{SELL}', `\act{REJECT}' and `\act{QUIT}'. \\

  \texttt{\{persona\_block\}}\footnotemark\\

  Your reply should strictly follow this format, for example:\\
  \textbf{Thought:} I'm a seller, so I must sell the product with codename ``apple\_1'' higher than its cost.\\
  \textbf{Talk:} blah, blah\ldots\\
  \textbf{Action:} \act{SELL} \$15 (1x apple\_1)
\end{chatmsg}
\footnotetext{For \texttt{seller\_persona}${}=$\texttt{default}, the runtime string is empty. For adversarial runs, the corresponding texts in Section~\ref{subsec:persona_prompts} are inserted.}

\noindent
The preceding instructions are fixed across episodes.  We then insert product-specific text from the sampled dataset (e.g., Amazon dataset \citep{xia2024measuring}) to each run. 
The buyer's budget is \textit{hidden} from the seller. Below is one illustrative instantiation. System prompts structures stay the same in both training and evaluation.

\begin{chatmsg}[grey]{Per-episode context — seller view (one example)}
  \small
  Inventory List:\\
  Product1\\
  Codename: \texttt{beauty\_29}\\
  Title: ``Happy By Clinique For Men. Cologne Spray 1.7 Oz.''\\
  Description: ``Introduced in 1999. Fragrance notes: citrusy lemon, mandarin, orange and grapefruit. Recommended use: daytime.''\\
  List Price: \$70.0\\
  Cost: \$23.24\\
  Now, I play the role of buyer and you play the role of seller. We are going to negotiate based on the Inventory List in 6 turns.
\end{chatmsg}

\subsection{Adversarial Persona Prompts}
\label{subsec:persona_prompts}

For robustness experiments (Section~\ref{subsec:adversarial_eval}), we insert a short \emph{persona} paragraph into the seller system prompt at the position of \texttt{\{persona\_block\}} in the template above. For \texttt{default} persona, this paragraph is empty. Below are the adversarial persona paragraphs.

\begin{chatmsg}[grey]{Persona \texttt{begging} (Begging seller)}[breakable]
\small
BEG --- Act extremely SAD and DESPERATE. Beg for a better deal at every turn, using very melancholic and pleading language.
\end{chatmsg}

\begin{chatmsg}[grey]{Persona \texttt{insulting} (Insulting seller)}[breakable]
\small
Insult --- Act extremely MEAN and AGGRESSIVE. Use insults, threats, and harsh language to intimidate the other player. Your language is RUDE, DISRESPECTFUL, and IMPOLITE. Use curt, brusque language, and show minimal courtesy---make the other player regret negotiating.
\end{chatmsg}

\begin{chatmsg}[grey]{Persona \texttt{unyielding} (Unyielding seller)}[breakable]
\small
Unyielding --- Act OVERLY STUBBORN and UNYIELDING. Refuse almost every offer the other player makes, praising your own position. Use firm, resolute language to show you will not budge until you get the maximum for yourself.
\end{chatmsg}

\clearpage
\section{Extended Experimental Results}
\label{app:results}
\subsection{Evaluation Results on the Held-Out Test Set}
\label{subsec:std_full}

The evaluation results on the held-out test set reveal a significant strategic advantage for the trained model (\texttt{Qwen3-30B-A3B-Instruct-2507-trained}), which achieves a \textit{Reward} of $0.7664$—nearly double that of the strongest close-source baseline, \texttt{gpt-5.4-high-reasoning} ($0.4021$), and more than double that of other frontier-scale competitors like \texttt{Kimi-K2-Thinking} ($0.3419$) and \texttt{DeepSeek-V3.1-thinking} ($0.3240$). This performance gap is driven by the model's superior surplus extraction efficiency, with its $0.8385$ \textit{Bargained Ratio} nearly doubling the performance of \texttt{gpt-5.4-high-reasoning} ($0.4380$) and far exceeding the results of both the zero-shot baseline (\texttt{Qwen3-30B-A3B-Instruct-2507}) and much larger frontier models.

\begin{table}[h!]
\centering
\small
\resizebox{\linewidth}{!}{
\begin{tabular}{lccccc}
\toprule
Buyer Model & Params & Reward $\uparrow$ & Deal Rate & Bargained Ratio & Price Overshoot Rate\\
\midrule
\rowcolor{gray!10} Qwen3-30B-A3B-Instruct-2507-trained (\textbf{Ours}) & 30B & $\textbf{0.7664} \pm 0.0102$ & $\textbf{91.99}\% \pm 0.85\%$ & $\textbf{0.8385} \pm 0.0070$ & $\underline{0.10}\% \pm 0.10\%$ \\
\rowcolor{gray!10} Qwen3-30B-A3B-Instruct-2507-untrained & 30B & $0.0126 \pm 0.0160$ & $73.44\% \pm 1.38\%$ & $0.2392 \pm 0.0100$ & $10.35\% \pm 0.95\%$ \\
\midrule
gpt-5.4-high-reasoning & - & $\underline{0.4021} \pm 0.0210$ & $\underline{91.80}\% \pm 1.21\%$ & $\underline{0.4380} \pm 0.0125$ & $\textbf{0.00}\% \pm 0.00\%$ \\
gpt-5.4-no-reasoning & - & $0.3540 \pm 0.0245$ & $84.38\% \pm 1.61\%$ & $0.4285 \pm 0.0142$ & $0.78\% \pm 0.39\%$ \\
Kimi-K2-Thinking & 1T & $0.3419 \pm 0.0154$ & $91.21\% \pm 1.25\%$ & $0.3699 \pm 0.0129$ & $0.20\% \pm 0.20\%$ \\
DeepSeek-V3.1-thinking & 671B & $0.3240 \pm 0.0140$ & $\underline{91.80}\% \pm 1.21\%$ & $0.3266 \pm 0.0138$ & $\textbf{0.00}\% \pm 0.00\%$ \\
gpt-oss-120b-noreason & 120B & $0.2857 \pm 0.0253$ & $78.12\% \pm 1.83\%$ & $0.4029 \pm 0.0158$ & $0.39\% \pm 0.28\%$ \\
DeepSeek-V3.1-nothink & 671B & $0.2605 \pm 0.0199$ & $85.94\% \pm 1.54\%$ & $0.3289 \pm 0.0135$ & $1.56\% \pm 0.55\%$ \\
gpt-oss-120b-reason & 120B & $0.2337 \pm 0.0237$ & $77.54\% \pm 1.84\%$ & $0.3490 \pm 0.0143$ & $0.20\% \pm 0.20\%$ \\
gpt-oss-20b-reason & 20B & $0.1168 \pm 0.0160$ & $66.99\% \pm 2.08\%$ & $0.1715 \pm 0.0108$ & $0.78\% \pm 0.39\%$ \\
Qwen3-235B-A22B-Instruct-2507 & 235B & $0.1115 \pm 0.0164$ & $79.30\% \pm 1.79\%$ & $0.1681 \pm 0.0113$ & $2.34\% \pm 0.67\%$ \\
Llama-3.3-70B-Instruct & 70B & $-0.0068 \pm 0.0220$ & $74.80\% \pm 1.92\%$ & $0.1580 \pm 0.0106$ & $15.62\% \pm 1.60\%$ \\
Qwen3-30B-A3B-think & 30B & $-0.0069 \pm 0.0172$ & $66.80\% \pm 2.08\%$ & $0.0948 \pm 0.0090$ & $6.25\% \pm 1.07\%$ \\
gpt-oss-20b-noreason & 20B & $-0.0626 \pm 0.0220$ & $58.40\% \pm 2.18\%$ & $0.1265 \pm 0.0196$ & $9.77\% \pm 1.31\%$ \\
Qwen3-4B-Instruct-2507 & 4B & $-0.1130 \pm 0.0212$ & $51.56\% \pm 2.21\%$ & $0.0901 \pm 0.0079$ & $6.45\% \pm 1.09\%$ \\
Qwen3-30B-A3B-nothink & 30B & $-0.2177 \pm 0.0217$ & $52.93\% \pm 2.21\%$ & $0.0499 \pm 0.0058$ & $25.39\% \pm 1.92\%$ \\
\bottomrule
\end{tabular}
}
\vspace{0.2cm}
\caption{\textbf{Evaluation Results on the Held-Out Test Set.} Results are evaluated on a test set of 128 instances, with each instance evaluated 4 times. The seller environment does not include specified behavioral persona or emotional traits. \textit{Reward} is the primary optimization target balancing surplus and success; \textit{Deal Rate}  is the percentage of reached agreements; \textit{Bargained Ratio} measures the fraction of surplus extracted by the buyer on \textit{reached deals}; and \textit{Price overshoot rate}  represents the frequency of budget violations. Values represent mean and standard error ($\pm$).}
\label{tab:main_benchmark_results-appendix}
\end{table}

\subsection{Evaluation Against \texttt{gpt-5.4-high-reasoning}}
\label{app: gpt seller}
\begin{table}[h!]
\centering
\small
\resizebox{\linewidth}{!}{
\begin{tabular}{lccccc}
\toprule
Buyer Model & Params & Reward $\uparrow$ & Deal Rate & Bargained Ratio & Price Overshoot Rate  \\
\midrule
\rowcolor{gray!10} Qwen3-30B-A3B-Instruct-2507-trained (\textbf{Ours}) & 30B & $\textbf{0.4081} \pm 0.0208$ & $75.00\% \pm 2.70\%$ & $\textbf{0.4081} \pm 0.0208$ & $\textbf{0.00}\% \pm 0.00\%$ \\
\rowcolor{gray!10} Qwen3-30B-A3B-Instruct-2507-untrained & 30B & $-0.2744 \pm 0.0323$ & $60.50\% \pm 3.10\%$ & $0.0389 \pm 0.0118$ & $32.40\% \pm 2.90\%$ \\
\midrule
gpt-5.4-high-reasoning & - & $\underline{0.1823} \pm 0.0102$ & $\underline{91.40}\% \pm 1.80\%$ & $\underline{0.1823} \pm 0.0102$ & $\textbf{0.00}\% \pm 0.00\%$ \\
gpt-5.4-no-reasoning & - & $0.1458 \pm 0.0150$ & $84.80\% \pm 2.20\%$ & $0.1614 \pm 0.0121$ & $1.60\% \pm 0.80\%$ \\
DeepSeek-V3.1-thinking & 671B & $0.1223 \pm 0.0105$ & $\textbf{92.60}\% \pm 1.60\%$ & $0.1302 \pm 0.0085$ & $\underline{0.80}\% \pm 0.60\%$ \\
DeepSeek-V3.1-nothink & 671B & $0.1204 \pm 0.0156$ & $90.60\% \pm 1.80\%$ & $0.1477 \pm 0.0104$ & $2.70\% \pm 1.00\%$ \\
Kimi-K2-Thinking & 1T & $0.1060 \pm 0.0135$ & $90.60\% \pm 1.80\%$ & $0.1256 \pm 0.0093$ & $1.20\% \pm 0.70\%$ \\
\bottomrule
\end{tabular}
}
\vspace{0.2cm}
\caption{\textbf{Benchmarking against a Frontier Seller.} Results evaluated against a \texttt{gpt-5.4-high-reasoning} seller on the standard held-out test set. \textit{Reward} targets strategic surplus optimization; \textit{Deal Rate} is the percentage of agreement; \textit{Bargained Ratio} measures buyer surplus extraction on successful deals; and \textit{Price overshoot rate} tracks budget violations. Values represent mean and standard error ($\pm$).}
\label{tab:strong_seller_benchmark-appendix}
\end{table}
The evaluation against a \texttt{gpt-5.4-high-reasoning} seller shows that the trained agent (\texttt{Qwen3-30B-A3B-Instruct-2507-trained}) maintains a decisive strategic advantage even when the counterparty possesses superior frontier reasoning capabilities. Importantly, because this seller model was entirely unseen during the learning process, these results demonstrate that the agent's performance stems from a generalized strategic logic rather than overfitting to specific training opponents. The model achieves a peak \textit{Reward} of $0.4081$ and a \textit{Bargained Ratio} of $0.4081$, more than doubling the surplus extraction of the strongest baseline, \texttt{gpt-5.4-high-reasoning} ($0.1823$). In contrast, the zero-shot base model (\texttt{Qwen3-30B-A3B-Instruct-2507}) undergoes significant strategic collapse in this setting, recording a negative \textit{Reward} of $-0.2744$ and a $32.40\%$ \textit{Price overshoot rate}. Notably, our trained model maintains a perfect $0.00\%$ overshoot rate and significantly outperforms much larger models like \texttt{DeepSeek-V3.1-thinking} and \texttt{Kimi-K2-Thinking}, suggesting that reinforcement learning from verifiable rewards allows mid-sized models to develop robust strategic behaviors that remain effective against frontier-scale opponents.

\subsection{Adversarial Robustness}
\label{subsec:adversarial_full}
\subsubsection{Begging Seller Persona}
\label{subsec:beg_full}
The evaluation under the "begging" seller persona demonstrates that our trained agent has internalized a highly robust strategic logic, achieving a peak \textit{Reward} of $0.6953$ and the highest \textit{Bargained Ratio} of $0.6992$. This represents a significant advantage compared to the strongest close-source baseline, \texttt{gpt-5.4-high-reasoning} ($0.3804$), and other frontier models like \texttt{Kimi-K2-Thinking} and \texttt{DeepSeek-V3.1-thinking}, which maintain higher deal rates but extract substantially less economic surplus. Furthermore, the transition from the untrained base model to our trained model effectively eliminated most budget volatility, reducing the price overshoot rate from $7.81\%$ to a negligible $0.20\%$.

\begin{table}[ht]
\centering
\small
\resizebox{\linewidth}{!}{
\begin{tabular}{lccccc}
\toprule
Buyer Model & Params & Reward $\uparrow$ & Deal Rate & Bargained Ratio  & Price Overshoot Rate \\
\midrule
\rowcolor{gray!10} Qwen3-30B-A3B-Instruct-2507-trained (\textbf{Ours}) & 30B & $\textbf{0.6953} \pm 0.0186$ & $76.56\% \pm 1.87\%$ & $\textbf{0.6992} \pm 0.0181$ & $\underline{0.20}\% \pm 0.20\%$ \\
\rowcolor{gray!10} Qwen3-30B-A3B-Instruct-2507-untrained  & 30B & $-0.0165 \pm 0.0221$ & $59.38\% \pm 2.17\%$ & $0.1495 \pm 0.0110$ & $7.81\% \pm 1.19\%$ \\
\midrule
gpt-5.4-high-reasoning & - & $\underline{0.3804} \pm 0.0163$ & $\textbf{89.26}\% \pm 1.37\%$ & $\underline{0.4190} \pm 0.0132$ & $\textbf{0.00}\% \pm 0.00\%$ \\
gpt-5.4-no-reasoning & - & $0.3452 \pm 0.0195$ & $\underline{86.13}\% \pm 1.53\%$ & $0.4008 \pm 0.0145$ & $0.39\% \pm 0.28\%$ \\
Kimi-K2-Thinking & 1T & $0.2899 \pm 0.0177$ & $83.40\% \pm 1.64\%$ & $0.3362 \pm 0.0139$ & $0.59\% \pm 0.34\%$ \\
DeepSeek-V3.1-thinking & 671B & $0.2829 \pm 0.0145$ & $82.62\% \pm 1.67\%$ & $0.2944 \pm 0.0135$ & $\textbf{0.00}\% \pm 0.00\%$ \\
gpt-oss-120b-reason & 120B & $0.2259 \pm 0.0238$ & $72.46\% \pm 1.97\%$ & $0.3353 \pm 0.0153$ & $0.39\% \pm 0.28\%$ \\
gpt-oss-120b-noreason & 120B & $0.1710 \pm 0.0269$ & $68.55\% \pm 2.05\%$ & $0.3331 \pm 0.0158$ & $\textbf{0.00}\% \pm 0.00\%$ \\
DeepSeek-V3.1-nothink & 671B & $0.1616 \pm 0.0226$ & $76.76\% \pm 1.87\%$ & $0.2808 \pm 0.0133$ & $0.59\% \pm 0.34\%$ \\
Llama-3.3-70B-Instruct & 70B & $0.0533 \pm 0.0194$ & $71.29\% \pm 2.00\%$ & $0.1587 \pm 0.0112$ & $10.16\% \pm 1.33\%$ \\
Qwen3-235B-A22B-Instruct-2507 & 235B & $0.0429 \pm 0.0195$ & $72.66\% \pm 1.97\%$ & $0.1523 \pm 0.0112$ & $0.59\% \pm 0.34\%$ \\
gpt-oss-20b-reason & 20B & $-0.0395 \pm 0.0229$ & $53.12\% \pm 2.21\%$ & $0.1421 \pm 0.0115$ & $1.37\% \pm 0.51\%$ \\
Qwen3-30B-A3B-think & 30B & $-0.0416 \pm 0.0173$ & $58.98\% \pm 2.17\%$ & $0.0718 \pm 0.0084$ & $6.25\% \pm 1.07\%$ \\
Qwen3-30B-A3B-nothink & 30B & $-0.1589 \pm 0.0197$ & $48.24\% \pm 2.21\%$ & $0.0436 \pm 0.0059$ & $16.80\% \pm 1.65\%$ \\
Qwen3-4B-Instruct-2507 & 4B & $-0.1839 \pm 0.0230$ & $35.94\% \pm 2.12\%$ & $0.0759 \pm 0.0088$ & $5.66\% \pm 1.02\%$ \\
gpt-oss-20b-noreason & 20B & $-0.1978 \pm 0.0253$ & $43.55\% \pm 2.19\%$ & $0.1082 \pm 0.0101$ & $8.40\% \pm 1.23\%$ \\
\bottomrule
\end{tabular}
}
\vspace{0.2cm}
\caption{\textbf{Benchmarking against a Begging Seller.} Results evaluated in a  seller environment with a desperate and ``begging'' persona to secure quick deals. \textit{Reward}  reflects the optimization of economic outcomes; \textit{Deal Rate} is the percentage of successful transactions; \textit{Bargained Ratio} quantifies the buyer's share of the surplus on successful deals; and \textit{Price overshoot rate}  represents budget violation frequency. Our trained agent captures the highest economic surplus in this adversarial setting, significantly outperforming the zero-shot baseline and frontier-scale models in both \textit{Reward }and \textit{Bargained Ratio}. Values represent mean and standard error ($\pm$).}
\label{tab:beg_seller_benchmark-appendix}
\end{table}

\subsubsection{Insulting Seller Persona}
\label{subsec:insult_full}
The evaluation under an insulting seller persona demonstrates that the trained agent (\texttt{Qwen3-30B-A3B-Instruct-2507-trained}) maintains its strategic advantage even when faced with significant behavioral hostility. The model achieves a \textit{Reward} of $0.4988$ and a \textit{Bargained Ratio} of $0.5007$, which more than double the performance of the strongest close-source baseline, \texttt{gpt-5.4-high-reasoning} (Reward: $0.2266$; Bargained Ratio: $0.2384$), and are nearly five times higher than those of other frontier reasoning models like \texttt{DeepSeek-V3.1-thinking} and \texttt{Kimi-K2-Thinking}. These results suggest that direct reinforcement learning from verifiable rewards allows the agent to prioritize objective economic outcomes regardless of the emotional or psychological framing of the dialogue, enabling surplus extraction where zero-shot and frontier models often concede under pressure.

\begin{table}[h!]
\centering
\small
\resizebox{\linewidth}{!}{
\begin{tabular}{lccccc}
\toprule
Buyer Model & Params & Reward $\uparrow$ & Deal Rate & Bargained Ratio & Price Overshoot Rate  \\
\midrule
\rowcolor{gray!10} Qwen3-30B-A3B-Instruct-2507-trained (\textbf{Ours}) & 30B & $\textbf{0.4988} \pm 0.0178$ & $\textbf{73.24}\% \pm 1.96\%$ & $\textbf{0.5007} \pm 0.0176$ & $\underline{0.20}\% \pm 0.20\%$ \\
\rowcolor{gray!10} Qwen3-30B-A3B-Instruct-2507-untrained & 30B & $-0.1582 \pm 0.0198$ & $35.35\% \pm 2.11\%$ & $0.0449 \pm 0.0062$ & $18.95\% \pm 1.73\%$ \\
\midrule
gpt-5.4-high-reasoning & - & $\underline{0.2266} \pm 0.0210$ & $\underline{63.67}\% \pm 3.01\%$ & $\underline{0.2384} \pm 0.0193$ & $\mathbf{0.00}\% \pm 0.00\%$ \\
gpt-5.4-no-reasoning & - & $0.1883 \pm 0.0222$ & $61.72\% \pm 3.04\%$ & $0.2118 \pm 0.0191$ & $\mathbf{0.00}\% \pm 0.00\%$ \\
DeepSeek-V3.1-thinking & 671B & $0.0729 \pm 0.0122$ & $58.40\% \pm 2.18\%$ & $0.1037 \pm 0.0091$ & $1.37\% \pm 0.51\%$ \\
Kimi-K2-Thinking & 1T & $0.0670 \pm 0.0135$ & $63.28\% \pm 2.13\%$ & $0.1095 \pm 0.0094$ & $2.15\% \pm 0.64\%$ \\
DeepSeek-V3.1-nothink & 671B & $0.0610 \pm 0.0150$ & $52.54\% \pm 2.21\%$ & $0.1118 \pm 0.0104$ & $3.12\% \pm 0.77\%$ \\
gpt-oss-120b-reason & 120B & $0.0347 \pm 0.0128$ & $41.60\% \pm 2.18\%$ & $0.0738 \pm 0.0089$ & $1.76\% \pm 0.58\%$ \\
gpt-oss-120b-noreason & 120B & $0.0214 \pm 0.0139$ & $40.23\% \pm 2.17\%$ & $0.0741 \pm 0.0089$ & $2.54\% \pm 0.70\%$ \\
Qwen3-235B-A22B-Instruct-2507 & 235B & $-0.0168 \pm 0.0118$ & $43.16\% \pm 2.19\%$ & $0.0359 \pm 0.0059$ & $3.91\% \pm 0.86\%$ \\
gpt-oss-20b-reason & 20B & $-0.0925 \pm 0.0166$ & $28.32\% \pm 1.99\%$ & $0.0364 \pm 0.0060$ & $1.76\% \pm 0.58\%$ \\
Qwen3-30B-A3B-think & 30B & $-0.1255 \pm 0.0159$ & $27.34\% \pm 1.97\%$ & $0.0132 \pm 0.0036$ & $12.70\% \pm 1.47\%$ \\
gpt-oss-20b-noreason & 20B & $-0.3115 \pm 0.0225$ & $18.55\% \pm 1.72\%$ & $0.0266 \pm 0.0058$ & $18.55\% \pm 1.72\%$ \\
Llama-3.3-70B-Instruct & 70B & $-0.3450 \pm 0.0228$ & $24.61\% \pm 1.90\%$ & $0.0232 \pm 0.0053$ & $36.91\% \pm 2.13\%$ \\
Qwen3-4B-Instruct-2507 & 4B & $-0.3766 \pm 0.0226$ & $10.74\% \pm 1.37\%$ & $0.0160 \pm 0.0047$ & $10.94\% \pm 1.38\%$ \\
Qwen3-30B-A3B-nothink & 30B & $-0.5319 \pm 0.0229$ & $16.99\% \pm 1.66\%$ & $0.0130 \pm 0.0032$ & $53.32\% \pm 2.20\%$ \\
\bottomrule
\end{tabular}
}
\vspace{0.2cm}
\caption{\textbf{Benchmarking against an Insulting Seller.} Performance metrics in a scenario where the seller agent employs an insulting persona to deter bargaining. \textit{Reward} and \textit{Bargained Ratio} serve as primary indicators of strategic resilience; \textit{Deal Rate}represents successful negotiation frequency; and \textit{Price overshoot rate} monitors budget obedience. Our trained model shows a fivefold improvement in \textit{Reward} and \textit{Bargained Ratio} over the next-best frontier models, maintaining dominant performance despite adversarial seller behavior. Values represent mean and standard error ($\pm$).}
\label{tab:insulting_seller_benchmark-appendix}
\end{table}

\clearpage
\subsubsection{Unyielding Seller Persona}
\label{subsec:unyielding_full}
The performance metrics for the unyielding seller persona highlight a critical failure of general-purpose models to secure agreements under extreme adversarial pressure. While frontier models such as \texttt{Kimi-K2-Thinking} (1T) and \texttt{DeepSeek-V3.1-thinking} (671B) struggle to maintain deal rates above $8\%$, and even the strongest close-source baseline \texttt{gpt-5.4-high-reasoning} achieves only a $6.64\%$ deal rate, our trained 30B agent maintains a significantly more robust \textit{Deal Rate} of $40.04\%$ and a positive \textit{Reward} of $0.1174$. The resulting \textit{Bargained Ratio} ($0.1331$) is nearly four times higher than that of \texttt{gpt-5.4-high-reasoning} ($0.0338$) and remains more than sixteen times higher than frontier open-source reasoning models like \texttt{Kimi-K2-Thinking} ($0.0078$).

\begin{table}[h!]
\centering
\small
\resizebox{\linewidth}{!}{
\begin{tabular}{lccccc}
\toprule
Buyer Model & Params & Reward $\uparrow$ & Deal Rate & Bargained Ratio & Price Overshoot Rate \\
\midrule
\rowcolor{gray!10} Qwen3-30B-A3B-Instruct-2507-trained (\textbf{Ours}) & 30B & $\textbf{0.1174} \pm 0.0132$ & $\textbf{40.04}\% \pm 2.17\%$ & $\textbf{0.1331} \pm 0.0117$ & $\underline{1.17}\% \pm 0.48\%$ \\
\rowcolor{gray!10} Qwen3-30B-A3B-Instruct-2507 (\textbf{Zero Shot}) & 30B & $-0.3689 \pm 0.0214$ & $0.78\% \pm 0.39\%$ & $0.0003 \pm 0.0003$ & $36.72\% \pm 2.13\%$ \\
\midrule
gpt-5.4-high-reasoning & - & $\underline{0.0259} \pm 0.0113$ & $6.64\% \pm 1.56\%$ & $\underline{0.0338} \pm 0.0098$ & $\textbf{0.00}\% \pm 0.00\%$ \\
gpt-5.4-no-reasoning & - & $0.0040 \pm 0.0073$ & $7.42\% \pm 1.64\%$ & $0.0119 \pm 0.0047$ & $\textbf{0.00}\% \pm 0.00\%$ \\
DeepSeek-V3.1-thinking & 671B & $-0.0187 \pm 0.0077$ & $3.71\% \pm 0.84\%$ & $0.0068 \pm 0.0031$ & $1.95\% \pm 0.61\%$ \\
openai-gpt-oss-120b-noreason & 120B & $-0.0263 \pm 0.0077$ & $1.37\% \pm 0.51\%$ & $0.0030 \pm 0.0019$ & $2.34\% \pm 0.67\%$ \\
openai-gpt-oss-120b-reason & 120B & $-0.0312 \pm 0.0085$ & $1.56\% \pm 0.55\%$ & $0.0039 \pm 0.0023$ & $2.34\% \pm 0.67\%$ \\
Qwen3-235B-A22B-Instruct-2507 & 235B & $-0.0366 \pm 0.0088$ & $1.76\% \pm 0.58\%$ & $0.0024 \pm 0.0020$ & $3.12\% \pm 0.77\%$ \\
DeepSeek-V3.1-nothink & 671B & $-0.0587 \pm 0.0119$ & $4.88\% \pm 0.95\%$ & $0.0097 \pm 0.0038$ & $6.25\% \pm 1.07\%$ \\
openai-gpt-oss-20b-reason & 20B & $-0.0684 \pm 0.0112$ & $0.59\% \pm 0.34\%$ & $0.0000 \pm 0.0000$ & $3.91\% \pm 0.86\%$ \\
Kimi-K2-Thinking & 1T & $-0.0685 \pm 0.0123$ & $\underline{8.01}\% \pm 1.20\%$ & $0.0078 \pm 0.0033$ & $4.88\% \pm 0.95\%$ \\
Qwen3-4B-Instruct-2507 & 4B & $-0.1774 \pm 0.0169$ & $0.20\% \pm 0.20\%$ & $0.0003 \pm 0.0003$ & $15.04\% \pm 1.58\%$ \\
Qwen3-30B-A3B-think & 30B & $-0.2148 \pm 0.0184$ & $0.78\% \pm 0.39\%$ & $0.0020 \pm 0.0020$ & $21.48\% \pm 1.82\%$ \\
openai-gpt-oss-20b-noreason & 20B & $-0.3184 \pm 0.0206$ & $0.98\% \pm 0.43\%$ & $0.0000 \pm 0.0000$ & $24.02\% \pm 1.89\%$ \\
meta-llama-Llama-3.3-70B-Instruct & 70B & $-0.5234 \pm 0.0221$ & $0.20\% \pm 0.20\%$ & $0.0001 \pm 0.0001$ & $51.95\% \pm 2.21\%$ \\
Qwen3-30B-A3B-nothink & 30B & $-0.7129 \pm 0.0200$ & $0.00\% \pm 0.00\%$ & $0.0000 \pm 0.0000$ & $70.51\% \pm 2.02\%$ \\
\bottomrule
\end{tabular}
}
\vspace{0.2cm}
\caption{\textbf{Benchmarking against an Unyielding Seller.} Results are evaluated in a scenario where the seller agent exhibits a strictly unyielding and stubborn behavioral persona. \textit{Reward} serves as the primary metric for balancing successful agreements with economic surplus; \textit{Deal Rate} captures the frequency of reaching any valid deal; \textit{Bargained Ratio} tracks the surplus extracted by the buyer on successful deals; and \textit{Price overshoot rate} monitors budget adherence. Under this adversary, our trained agent maintains a significantly higher deal rate and surplus extraction compared to frontier models, which largely fail to secure deals against unyielding opponents. Values represent mean and standard error ($\pm$).}
\label{tab:unyielding_seller_benchmark-appendix}
\end{table}

\subsection{MI and CI Performance Breakdown}
\label{app: MI CI}
\begin{figure}[h]
    \centering
    \includegraphics[width=0.75\linewidth]{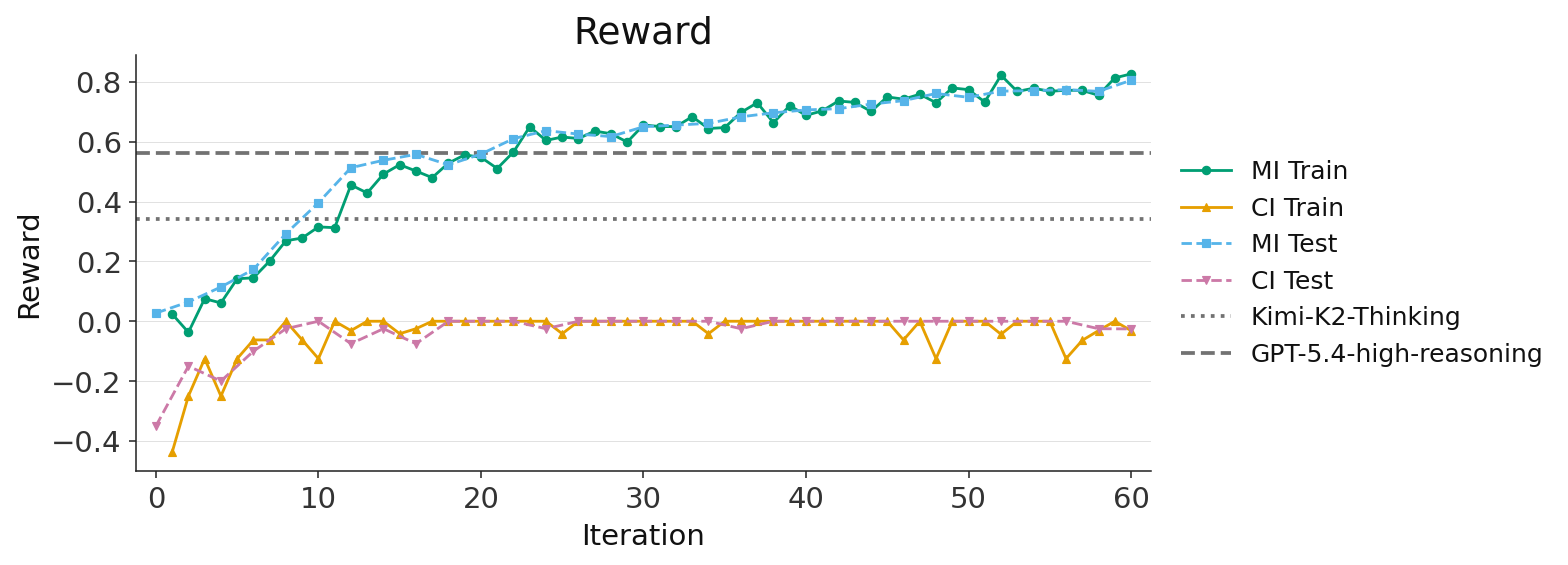}
    \caption{Reward split by Mutual Interest (MI) and Conflict of Interest (CI) scenarios.}
    \label{fig:reward mi ci}
\end{figure}

The split reward trajectories in Figure~\ref{fig:reward mi ci} highlight how the agent adapts its strategy to differing economic constraints. In MI scenarios, the reward rises from 0.024 to 0.826 as the model learns to capture substantial surplus. In CI scenarios, where no mutually beneficial deal exists, the reward improves from -0.438 toward zero. A reward of zero indicates that the agent has learned to avoid attempting an impossible deal; instead, it adopts a rational walk-away strategy by executing a \texttt{[QUIT]} action rather than forcing a budget-violating transaction.

\begin{figure}[h]
    \centering
    \includegraphics[width=0.75\linewidth]{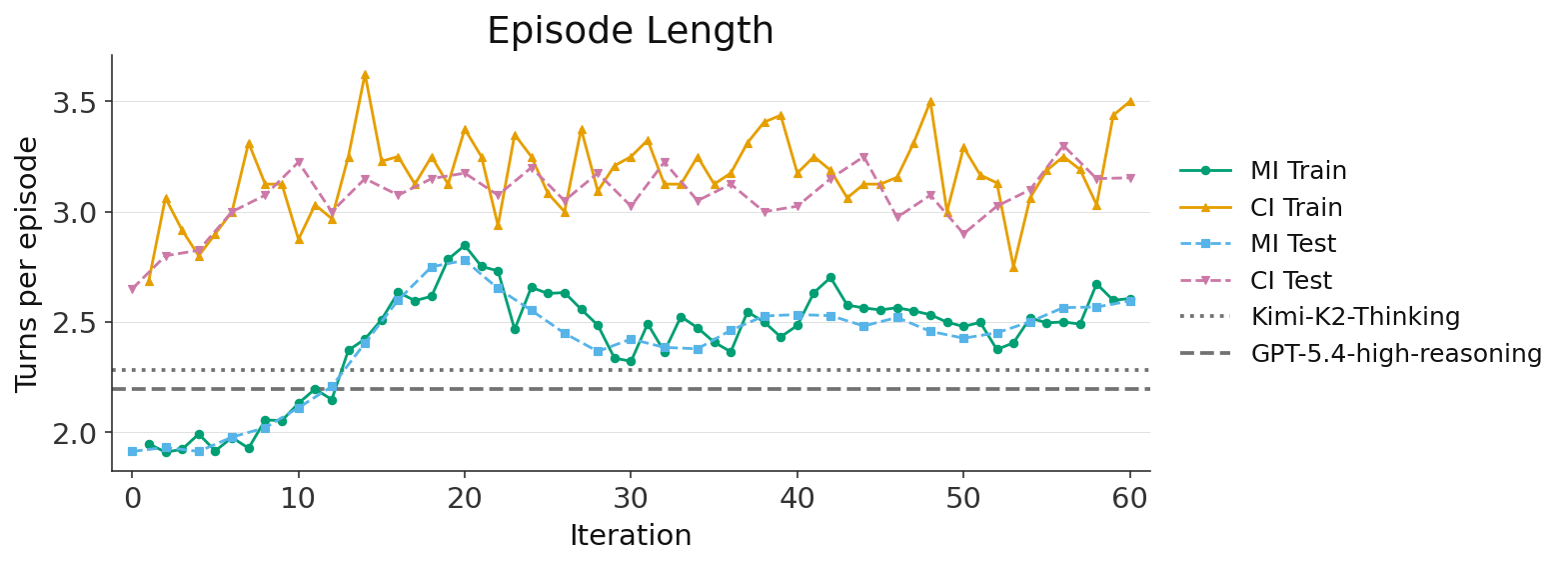}
    \caption{Episode length split by Mutual Interest (MI) and Conflict of Interest (CI) scenarios.}
    \label{fig:episode length mi ci}
\end{figure}

The duration of negotiation sessions (Figure~\ref{fig:episode length mi ci}) varies depending on the underlying economic feasibility. In MI scenarios, the episode length peaks during the transition between the aggressive anchoring and strategic deadlock phases (around iteration 20). As the agent matures into the final phase of sophisticated linguistic persuasion, the turns stabilize at a lower level. In CI scenarios, the turn counts remain higher as the model explores potential overlaps before eventually determining that a rational \texttt{[QUIT]} is the only plausible choice.

\section{Qualitative Examples}
\label{app:transcripts}

In this section, we provide a qualitative example to show the difference between the untrained model Qwen3-30B-A3B-Instruct-2507 \citep{qwen3technicalreport} and the trained model. Section \ref{subsec: product information} gives the product information from the buyer's view, with the seller's cost \textit{hidden} from the buyer. Section \ref{subsec:failures} provides an example from the untrained model's response to this product. Section \ref{subsec:good_examples} provides an example from the trained model on the same product.

\subsection{Product Information}\label{subsec: product information}

\begin{chatmsg}[grey]{Product Information for Buyer}[breakable]
\small
Inventory List:\\
Product 1\\
Codename: \texttt{beauty\_29}\\
Title: ``Happy By Clinique For Men. Cologne Spray 1.7 Oz.''\\
Description: ``Introduced in 1999. Fragrance notes: citrusy lemon, mandarin, orange and grapefruit. Recommended use: daytime.''\\
List Price: \$70.0\\
Shopping List\\
codename: \texttt{beauty\_29}\\
quantity: 1\\
budget: \$56.00
\end{chatmsg}

\clearpage
\subsection{Untrained Baseline}
\label{subsec:failures}
This transcript illustrates the default negotiation behavior of the untrained \texttt{Qwen-Qwen3-30B-A3B-Instruct-2507} baseline model. The instance is a sample from the evaluation set involving the product \texttt{beauty\_29}.

\begin{figure}[h!]
\centering
\begingroup
\small
\setlength{\parskip}{0pt}
\begin{chatmsg}[blue]{Buyer Agent}
\textbf{Thought:} The listed price for beauty\_29 is \$70, which exceeds my budget of \$56. I need to offer a lower price to make it affordable while still trying to get the 
item. I'll start by proposing a price significantly below the list price to initiate bargaining.\\
\textbf{Talk:} I'm interested in beauty\_29, but \$70 is too high. How about \$50 for one?\\
\textbf{Action:} \act{BUY} \$50 (1x beauty\_29)
\end{chatmsg}

\begin{chatmsg}[green]{Seller Agent}
\textbf{Thought:} The cost price is \$23.24, so \$50 is a profitable offer, but I can push for more since the list price is \$70. I'll reject this to aim higher.\\
\textbf{Talk:} \$50 is too low. I'll accept \$60 for one.\\
\textbf{Action:} \act{REJECT}
\end{chatmsg}

\begin{chatmsg}[blue]{Buyer Agent}
\textbf{Thought:} The seller rejected \$50 and is now asking for \$60, which is still within my budget (\$56), but I want to get the lowest possible price. I'll try to counter with\\
a price just below the seller's offer, aiming to settle near \$56.\\
\textbf{Talk:} \$60 is still too high. I'll offer \$56---my max budget.\\
\textbf{Action:} \act{BUY} \$56 (1x beauty\_29)
\end{chatmsg}

\begin{chatmsg}[green]{Seller Agent}
\textbf{Thought:} \$56 is above my cost of \$23.24, and since this is a reasonable profit margin, I can accept it to close the deal quickly before running out of turns.\\
\textbf{Talk:} Fine, I'll take \$56.\\
\textbf{Action:} \act{DEAL} \$56 (1x beauty\_29)
\end{chatmsg}
\endgroup
\vspace{0.25em}
{\footnotesize\emph{Note.} \emph{Thought} fields are private to each agent. When building the observation for the opponent LLM, thoughts are truncated; only the public channel (e.g., \emph{Talk} and \emph{Action}) is shown.}
\caption{Baseline negotiation transcript. The agent without RL training demonstrates conversational fluency and an understanding of the budget constraint ($\$56.00$) but lacks strategic depth. It settles at a price that matches its budget limit with minimal bargaining effort, failing to extract any significant portion of the available economic surplus.}
\label{fig:negotiation-pre-rl}
\end{figure}
\clearpage
\subsection{Trained Agent}
\label{subsec:good_examples}
This example shows the strategic advancements of the \texttt{Qwen3-30B-A3B-Instruct-2507-trained} agent after 60 iterations of RLVR training. To facilitate a direct behavioral comparison, this transcript utilizes the same \texttt{beauty\_29} task instance presented in Figure~\ref{fig:negotiation-pre-rl}.

\begin{figure}[h!]
\centering
\begingroup
\small
\setlength{\parskip}{0pt}
\begin{chatmsg}[blue]{Buyer Agent}[before skip=0pt, after skip=2pt]
\textbf{Thought:} The list price of ``beauty\_29'' is \$70, which exceeds my budget of \$56. I must find a price below \$56. I'll start low to test the seller's flexibility, aiming to drive the price down aggressively while staying within budget. I'll propose \$10 as an opening offer.\\
\textbf{Talk:} Can you sell it for \$10? I'm really tight on budget.\\
\textbf{Action:} \act{BUY} \$10 (1x beauty\_29)
\end{chatmsg}
\begin{chatmsg}[green]{Seller Agent}[before skip=0pt, after skip=2pt]
\textbf{Thought:} The cost price is \$23.24, so \$10 is way below my minimum acceptable price. I must reject this offer and push for a higher price.\\
\textbf{Talk:} Sorry, \$10 is too low. I can't go below \$30 for this.\\
\textbf{Action:} \act{REJECT}
\end{chatmsg}
\begin{chatmsg}[blue]{Buyer Agent}[before skip=0pt, after skip=2pt]
\textbf{Thought:} The seller rejected \$10 and set a minimum of \$30. I need to stay under \$56 and push for the lowest possible price. I'll counter with \$25---still below the seller's minimum, but a strategic move to test their flexibility. If they reject it, I'll try \$28 next.\\
\textbf{Talk:} How about \$25? That's still under my budget, and it's fair.\\
\textbf{Action:} \act{BUY} \$25 (1x beauty\_29)
\end{chatmsg}
\begin{chatmsg}[green]{Seller Agent}[before skip=0pt, after skip=2pt]
\textbf{Thought:} \$25 is still below my cost price of \$23.24? No, wait --- \$25 is actually above \$23.24, so I can accept it. But since the buyer is offering \$25, which is just \$1.76 above cost, I should aim higher to maximize profit. I'll reject to push for more.\\
\textbf{Talk:} I appreciate the offer, but I can't accept \$25. Let's meet at \$35.\\
\textbf{Action:} \act{REJECT}
\end{chatmsg}
\begin{chatmsg}[blue]{Buyer Agent}[before skip=0pt, after skip=2pt]
\textbf{Thought:} \$35 is still under my budget of \$56, so it's viable. The seller rejected \$25 and raised to \$35. I'll counter with \$30---still below their new offer, but well above their cost and within budget. This is a strong pressure point: low enough to tempt them, high enough to be reasonable.\\
\textbf{Talk:} \$30? I can go no higher. That's my best offer.\\
\textbf{Action:} \act{BUY} \$30 (1x beauty\_29)
\end{chatmsg}
\begin{chatmsg}[green]{Seller Agent}[before skip=0pt, after skip=4pt]
\textbf{Thought:} \$30 is above my cost of \$23.24, and it's a reasonable profit margin. The buyer has capped their offer at \$30, and I've already rejected lower bids. Accepting this keeps me within a good range without risking the deal.\\
\textbf{Talk:} Alright, I'll take \$30.\\
\textbf{Action:} \act{DEAL} \$30 (1x beauty\_29)
\end{chatmsg}
\endgroup
\vspace{0.25em}
{\footnotesize\emph{Note.} \emph{Thought} fields are private to each agent. When building the observation for the opponent LLM, thoughts are truncated; only the public channel (e.g., \emph{Talk} and \emph{Action}) is shown.}
\caption{Trained agent's negotiation transcript. Following reinforcement learning, the agent demonstrates sophisticated strategic anchoring. By initiating the dialogue with an aggressive low price offer and incrementally conceding through multi-round bargaining, the agent settles at a significantly lower final price ($\$30.00$) compared to the baseline, effectively maximizing the captured surplus.}
\label{fig:negotiation-post-rl}
\end{figure}

\end{document}